\documentclass[]{gtech}

\usepackage{amssymb}
\usepackage{bigdelim}
\usepackage{todonotes}
\usepackage{longtable}
\usepackage{tabularray}
\usepackage[most]{tcolorbox} 
\usepackage{algorithm}  
\usepackage{algpseudocode} 
\usepackage[toc,page]{appendix}
\usepackage[normalem]{ulem}
\usepackage{adjustbox}
\usepackage{natbib}
\usepackage{colortbl}  
\usepackage{subcaption}
\usepackage{pifont}
\usepackage{amsmath}
\usepackage{amsfonts}       
\usepackage{mathrsfs}
\usepackage{tabularx}
\usepackage{hyperref}       
\usepackage{url}            
\usepackage{booktabs}       
\usepackage{nicefrac}       
\usepackage{microtype}      
\usepackage{xcolor}         
\definecolor{ourreward}{HTML}{32bbee}
\definecolor{sparsereward}{HTML}{ed9a81}
\usepackage{graphicx}
\usepackage{tikz}
\usepackage{fontawesome5}
\usepackage{wrapfig}
\usepackage{needspace}
\usepackage{cleveref}
\usepackage{fvextra}
\tcbuselibrary{listings, breakable}

\usepackage{bbding}
\usepackage{perpage}
\MakePerPage{footnote}
\usepackage{caption}
\usepackage{CJK}

\usepackage{bbm}
\useunder{\uline}{\ul}{}

\title{VisSearch-Bench: Evaluating Long Horizon Visual Search in Multimodal Agents}

\author[1,2,*,\ddagger]{ZhengXian~Wu}
\author[1,2,*]{Qianhao~Nie}
\author[1,*]{Zhuohong~Chen}
\author[2]{Junjie~Gao}
\author[2,3]{Yuyao~Zhang}
\author[1]{Yunyao~Yu}
\author[1]{Hangrui~Xu}
\author[2]{Cheng~Cheng}
\author[2]{Guiyu~Ma}
\author[2]{Jin~Zhang}
\author[2]{Xingzhong~Xu}
\author[2,\dagger]{Kai~Yang}
\author[1,\dagger]{Haoqian~Wang}

\affiliation[1]{Tsinghua University\quad}
\affiliation[2]{Ant Group\quad}
\affiliation[3]{Renmin University of China}

\contribution[*]{Equal contribution}
\contribution[\dagger]{Corresponding author}
\contribution[\ddagger]{Work done during an internship at Ant Group}

\usepackage[nointegrals]{wasysym}
\newfontfamily\vissearchresultfont{texgyrepagella-regular.otf}
\AtBeginDocument{\let\cite\citep}
\pretocmd{\section}{\ifnum\value{section}>0\FloatBarrier\fi}{}{}
\pretocmd{\subsection}{\FloatBarrier}{}{}
\definecolor{vissearchblue}{HTML}{0F5FA8}
\definecolor{vissearchgreen}{HTML}{246B3D}
\definecolor{vissearchamber}{HTML}{9A6700}
\definecolor{vissearchgray}{HTML}{9CA3AF}
\definecolor{vissearchrule}{HTML}{D1D5DB}
\definecolor{vissearchgroup}{HTML}{EAF2F7}
\definecolor{vissearchsubgroup}{HTML}{F4F7FA}
\DeclareRobustCommand{\vsfull}{\textcolor{vissearchgreen}{\ensuremath{\checkmark}}}
\DeclareRobustCommand{\vspartial}{\textcolor{vissearchamber}{\LEFTcircle}}
\DeclareRobustCommand{\vsnone}{\textcolor{vissearchgray}{\textemdash}}

\lstdefinestyle{vissearchprompt}{
  basicstyle=\ttfamily\scriptsize,
  breaklines=true,
  breakatwhitespace=true,
  columns=fullflexible,
  keepspaces=true,
  aboveskip=0pt,
  belowskip=0pt,
  showstringspaces=false
}

\newtcolorbox{actorpromptbox}[1]{
  colback=blue!6!white,
  colframe=blue!50!black,
  title=\textbf{#1},
  fonttitle=\bfseries,
  sharp corners,
  breakable
}

\newtcolorbox{judgepromptbox}[1]{
  colback=yellow!10!white,
  colframe=yellow!50!black,
  title=\textbf{#1},
  fonttitle=\bfseries,
  sharp corners,
  breakable
}

\newtcolorbox{formatpromptbox}[1]{
  colback=gray!5!white,
  colframe=gray!50!black,
  title=\textbf{#1},
  fonttitle=\bfseries,
  sharp corners,
  breakable
}

\abstract{
\vspace{-1em}
\begin{center}
\bfseries Abstract
\end{center}
\vspace{-0.5em}
Multimodal agents use web and image tools to answer questions beyond their
internal knowledge. Some search tasks use images mainly to identify an initial
entity, then rely on text retrieval. These tasks offer limited insight into how
agents connect information across images, objects, and search steps. We therefore
focus on \textbf{visual context management}. This involves organizing and using
visual information and linking it to other findings during search.
Visually intensive tasks require agents to
combine object appearance, local details, or spatial relations with external
information.
\setlength{\parskip}{0pt}
We introduce \textbf{VisSearch-Bench}, with 204 human-written tasks across
7 main domains and 25 subdomains. It covers Set-wise Visual Search, Sequential Visual Search,
Hybrid Wide--Deep Search, Competitive Visual Disambiguation, and
Visual-Structure-Guided Search. Each task provides a reference answer, traceable
evidence, and scoring rules. \textbf{Accuracy} measures complete task success,
while \textbf{Process Score} measures the correctness and completeness of
individual results and subquestion answers. Across 16 multimodal models, the
best system fully solves only 26.47\% of tasks, even with search tools. These
results highlight the challenge of linking visual evidence with search results
to produce complete answers.
\par
}

\gtechdata[\textcolor{gtechblue}{\faGlobe}\ Website]{\href{https://dingwu1021.github.io/VisSearch-Bench/}{Project page}}
\gtechdata[\faGithub\ Code]{\href{https://github.com/antins-labs/VisSearch-Bench-v1}{GitHub repository}}
\gtechdata[\textcolor{orange!80!black}{\faDatabase}\ Hugging Face]{\href{https://huggingface.co/datasets/antins-labs/VisSearch-Bench-v1}{Dataset}}

\begin{document}
\maketitle
\begin{figure}[H]
\centering
\includegraphics[width=\textwidth]{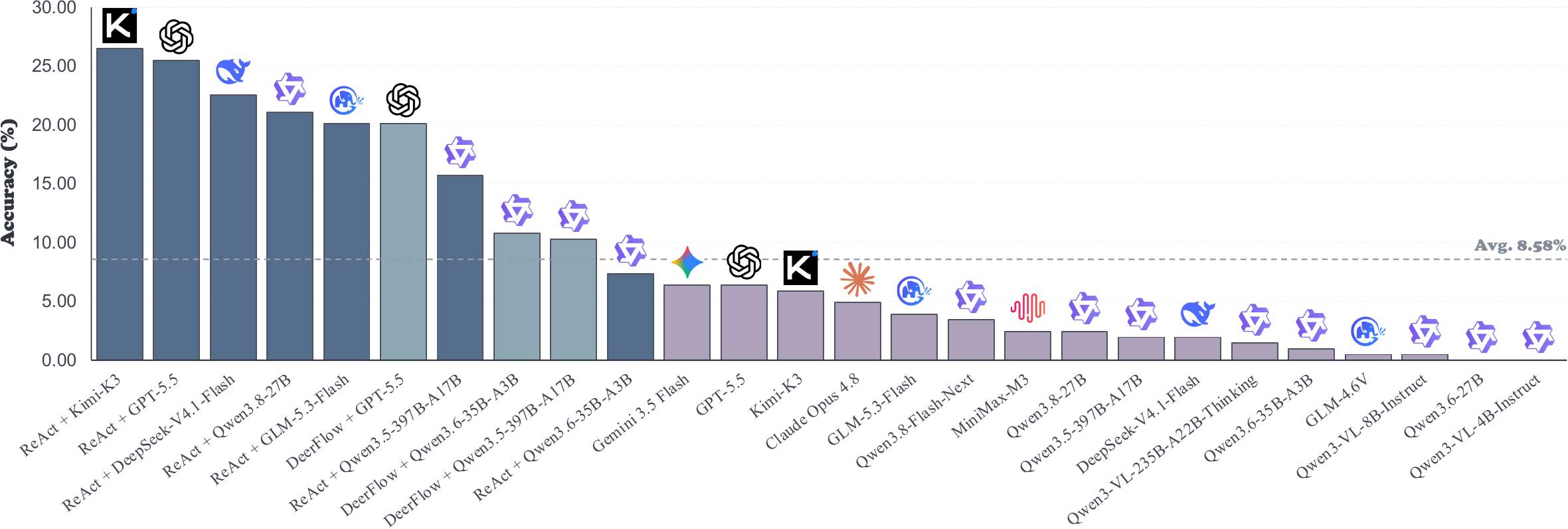}
\caption{Accuracy of models and agents on \textbf{VisSearch-Bench}, comparing direct answering with ReAct and DeerFlow. The dashed line marks the average accuracy across the systems shown.}
\label{fig:accuracy-overview}
\end{figure}

\section{Introduction}

Multimodal large language models (MLLMs) are evolving from one-shot visual question answering systems into agents that can plan, use tools, and interact with the open web over multiple rounds~\cite{Luo2026FromNT,wu2026promsa}. With text search, image search, webpage browsing, and region-level image inspection, these systems can gather evidence beyond the initial prompt and solve complex information-seeking tasks~\cite{Chng2025SenseNovaMARSEM, zhang2026searchos}. This paradigm is closer to real-world information seeking than visual question answering based only on parametric knowledge.

At the same time, multimodal benchmarks are placing stronger demands on visual information. Models must not only understand images in the prompt, but also combine visual and textual evidence found during search~\cite{Hou2026InterLVSearchBI}. MMSearch-Plus~\cite{Tao2025MMSearchPlusBP} emphasizes the extraction and propagation of fine-grained visual cues. MM-BrowseComp~\cite{Li2025MMBrowseCompAC} and VisBrowse-Bench~\cite{Zhang2026VisBrowseBenchBV} require agents to actively obtain visual evidence while browsing. 
Together, these benchmarks make visual evidence an active component of search.

However, existing multimodal search benchmarks often focus on whether visual evidence is used at some point in a task. How images are used can lead to different task demands. In some existing tasks, images mainly serve to identify an initial entity~\cite{Wang2023FillingTI,Chen2026LearningTS}. Once the agent obtains the name of a person, object, or place, it can rely mainly on text search. Such tasks provide limited insight into how agents handle multiple images or connect visual information across search steps. More recent benchmarks involve more image searches and tool calls. However, higher counts alone do not imply more complex visual context. We also need to examine how these visual operations are connected and how they support later searches and the final answer.

Real-world information seeking is more than a series of separate image recognition tasks~\cite{zhang2026searchos}. On the open web, agents need to update their understanding as they gather new evidence~\cite{Yao2022ReActSR}. Early evidence may suggest an initial hypothesis. Later evidence may help compare candidates, rule some out, or identify new ones. New information may also change how an earlier image is understood. An agent therefore needs to connect visual evidence across search steps and use it to guide further search~\cite{Hu2023AVISAV}. For example, finding all works in a catalog that meet a visual condition requires several steps. The agent must first establish the full set of entries, then inspect their images and combine the results. This raises a basic question: \textbf{Can agents reliably organize and use visual evidence as it accumulates during open-web search?}

We use \textbf{visual context management} to describe how agents organize and use visual information and its connections to findings during search. We study this ability through tasks that involve multiple images, multiple objects, and linked search steps. We examine whether agents can connect visual observations with external information to compare objects, filter candidates, and follow clues. We also examine whether they can combine results across steps into a complete and consistent answer.

\begin{figure}[!t]
\captionsetup{skip=4pt}
\centering
\includegraphics[width=\textwidth]{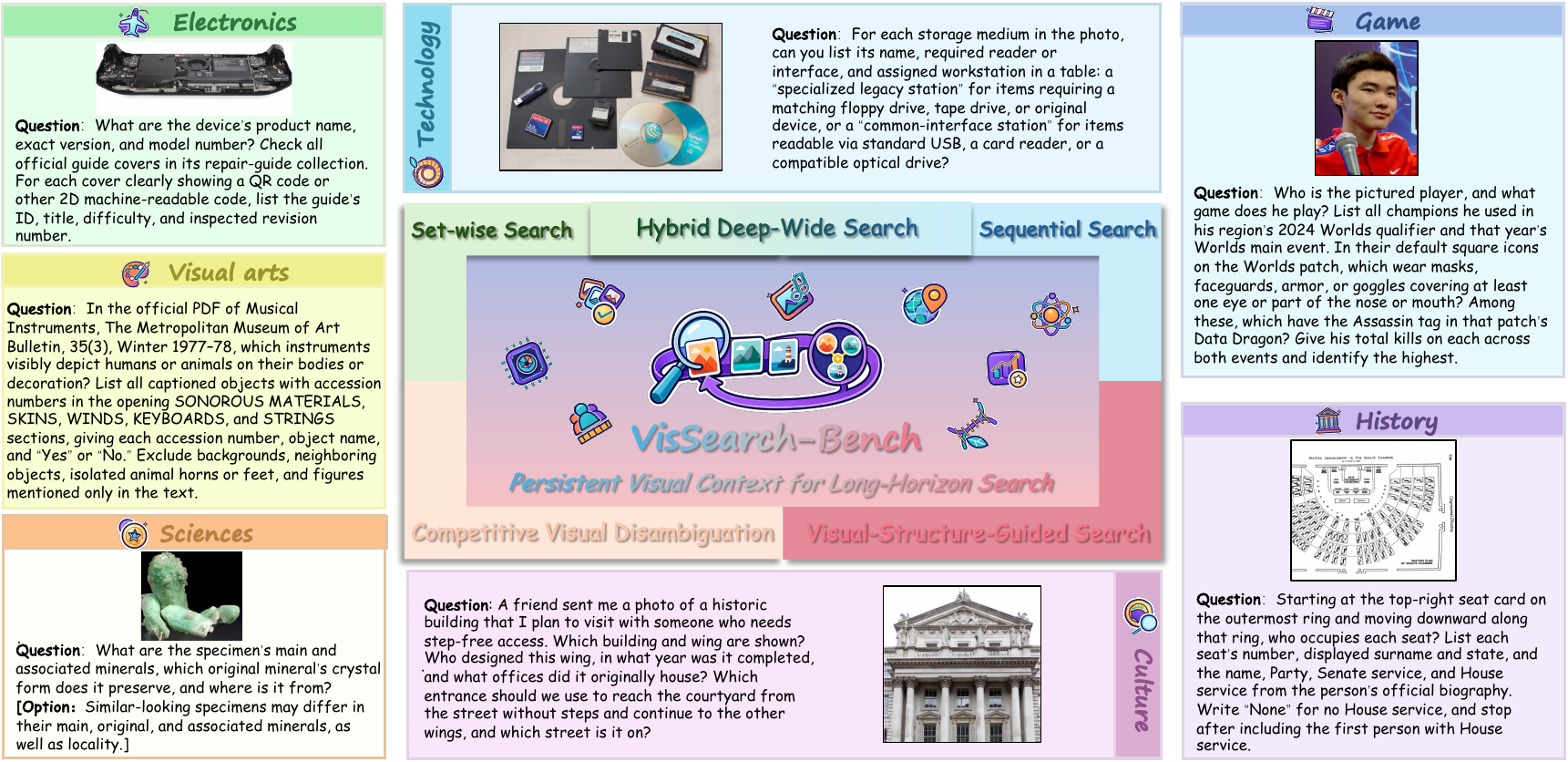}
\caption{Overview of VisSearch-Bench. The center presents five task families that require long-horizon visual search, while the surrounding panels show selected examples from the dataset across diverse content domains.}
\label{fig:vissearch-overview}
\vspace{-14pt}
\end{figure}

To evaluate this ability, we introduce \textbf{VisSearch-Bench}, with five task categories that reflect different demands of long-horizon visual search. \textbf{Set-wise Visual Search} requires agents to collect and check visual information for a set of objects, then filter, compare, or combine the results. \textbf{Sequential Visual Search} uses visual clues from earlier steps to guide later searches. \textbf{Hybrid Wide--Deep Search} combines step-by-step search with search over a set of objects. Results from one stage determine the search scope or targets of another. \textbf{Competitive Visual Disambiguation} focuses on using fine visual details to distinguish similar-looking candidates. \textbf{Visual-Structure-Guided Search} combines object order, spatial positions, or connections in an image with information found through external search. Agents use this information to locate objects, check conditions, and combine results. Together, these tasks examine how agents handle multiple images, connect search results, and reason across steps. Figure~\ref{fig:vissearch-overview} presents the five categories and representative tasks from different domains.

VisSearch-Bench contains 204 human-written tasks across 7 main domains and 25 subdomains. Dataset construction is led by human annotators and supported by automated checks. Annotators build each task from public, searchable sources. They gather the required images, entity relationships, and facts, then write the question and reference answer. They also document the evidence supporting the solution. Automated checks and human review examine source accessibility, answer consistency, and possible shortcuts that bypass the intended visual requirements. For evaluation, we use \textbf{Accuracy} to measure whether the full answer meets the task requirements. We also use a task-specific \textbf{Process Score} to assess the correctness and completeness of individual results and answers to subquestions. Our main contributions are as follows:

\begin{enumerate}
    \setlength{\parsep}{0pt}
    \setlength{\parskip}{0pt}
    \setlength{\itemsep}{1pt}
    \item We introduce a task-based perspective for evaluating \textbf{visual context management} in open-web search. It focuses on how agents organize information from multiple images, connect different objects, and use visual findings to guide later searches.
    \item We construct \textbf{VisSearch-Bench}, with 204 human-written tasks across 7 main domains and 25 subdomains. Its five categories cover Set-wise Visual Search, Sequential Visual Search, Hybrid Wide--Deep Search, Competitive Visual Disambiguation, and Visual-Structure-Guided Search.
    \item We develop a human-led construction process supported by automated checks. Each task includes evidence sources, a reference answer, and scoring rules. Our evaluation combines full-answer accuracy with scores for individual results and subquestions. This measures overall task completion while also capturing which parts of an answer are correct or missing.
\end{enumerate}

\section{Related Work}

\subsection{Vision-Language Models}

Vision-language models (VLMs) combine visual information with language understanding~\cite{Wu2026WhenMJ,Yu2026StabilizingUS}. Their capabilities have expanded from single-image recognition and question answering to document reading, multi-image understanding, video analysis, and visual reasoning. InternVL3~\cite{Zhu2025InternVL3EA} learns language and visual capabilities through joint pre-training and supports longer multimodal contexts. Kimi-VL~\cite{Du2025KimiVLTR} focuses on high-resolution images, multi-image understanding, and long-context reasoning. Qwen3-VL~\cite{Bai2025Qwen3VLTR} supports interleaved text, image, and video inputs, with improvements in multimodal reasoning and spatial understanding. These advances provide a foundation for handling more varied and complex visual inputs.

\begin{table}[!t]
\centering
\fontsize{10}{12.4}\selectfont
\setlength{\tabcolsep}{2pt}
\renewcommand{\arraystretch}{1.22}
\newcommand{\vsbenchhead}[1]{{\fontsize{9.5}{11.2}\selectfont\bfseries #1}}
\begin{tabularx}{\textwidth}{l*{7}{>{\centering\arraybackslash}X}}
\toprule
\vsbenchhead{Benchmarks}
& \vsbenchhead{\shortstack{Open-\\Web}}
& \vsbenchhead{\shortstack{Image\\Search}}
& \vsbenchhead{\shortstack{Multi-\\image}}
& \vsbenchhead{\shortstack{Multi-\\hop}}
& \vsbenchhead{\shortstack{Set\\Answers}}
& \vsbenchhead{\shortstack{Evidence\\Links}}
& \vsbenchhead{\shortstack{Partial\\Credit}} \\
\midrule
BrowseComp
& \vsfull & \vsnone & \vsnone & \vsfull
& \vsnone & \vsnone & \vsnone \\
MMSearch
& \vsfull & \vsfull & \vspartial & \vspartial
& \vsnone & \vsfull & \vspartial \\
BrowseComp-VL
& \vsfull & \vsfull & \vsnone & \vsfull
& \vsnone & \vsnone & \vsnone \\
MM-BrowseComp
& \vsfull & \vsfull & \vsfull & \vsfull
& \vsnone & \vsfull & \vsfull \\
MMSearch-Plus
& \vsfull & \vsfull & \vsfull & \vsfull
& \vsnone & \vspartial & \vsnone \\
VisBrowse-Bench
& \vsfull & \vsfull & \vsfull & \vsfull
& \vsnone & \vsnone & \vsnone \\
BrowseComp-V3
& \vsfull & \vsfull & \vsfull & \vsfull
& \vsnone & \vsfull & \vsfull \\
InterLV-Search
& \vsfull & \vsfull & \vsfull & \vsfull
& \vsnone & \vsnone & \vsnone \\
VDR-Bench
& \vsfull & \vsfull & \vsnone & \vsfull
& \vsnone & \vsnone & \vsnone \\
\midrule
\rowcolor[HTML]{E2F0D9}
\textbf{VisSearch-Bench (Ours)}
& \vsfull & \vsfull & \vsfull & \vsfull
& \vsfull & \vsfull & \vsfull \\
\bottomrule
\end{tabularx}
\caption{Comparison of representative open-web and multimodal search benchmarks. \vsfull\ Full; \vspartial\ Partial; \vsnone\ Not covered. InterLV-Search refers to its open-web Level~3.}
\label{tab:benchmark-comparison}
\end{table}

Recent models further combine visual understanding with general reasoning, tool use, and multi-step task execution. Qwen3.5~\cite{qwen3.5} uses unified vision-language training and reinforcement learning to improve reasoning and agent capabilities. Later models, including Qwen3.6-35B-A3B~\cite{qwen3.6-35b-a3b} and Qwen3.8-27B~\cite{qwen38}, support image and video understanding. They also improve code generation and multi-step task execution. Visual understanding is thus becoming an important part of general-purpose agent models. It supports tasks such as interpreting charts, reading documents, and using information from user interfaces.

Other model families follow a similar direction. GLM-5.3-Flash~\cite{glm5team2026glm5vibecodingagentic} introduces native multimodal capabilities to the GLM-5 series. It supports browser and computer use based on visual information. DeepSeek-V4.1-Flash~\cite{Xu2026DeepSeekV41FlashPT} supports both visual inputs and long-context reasoning. Kimi K3~\cite{Bai2026KimiKO} also provides native image and video understanding. Its one-million-token context window supports extended reasoning and tool interactions.

However, longer context windows do not guarantee effective visual-state management during search~\cite{wu2026promsa}. In open-web search, agents also need to correctly match objects across images and connect visual observations with external facts. They must use earlier findings to determine later search targets. VisSearch-Bench therefore focuses on visual context management in tasks involving multiple images, multiple objects, and linked search steps. It evaluates whether agents can use these connections to verify objects, compare candidates, filter results, and combine findings across steps.

\subsection{Agentic Information Seeking}

Existing benchmarks evaluate agentic information seeking through reasoning, tool use, and information gathering. GAIA focuses on multi-step reasoning and tool use in realistic tasks~\cite{Mialon2023GAIAAB}. BrowseComp requires agents to persistently search the open web for facts that are difficult to find but easy to verify~\cite{Wei2025BrowseCompAS}. WideSearch extends the focus to large-scale information gathering, emphasizing complete results that can be checked item by item~\cite{Wong2025WideSearchBA}. DeepWideSearch combines multi-hop retrieval with broad information collection to evaluate whether agents can handle both search depth and breadth~\cite{Lan2025DeepWideSearchBD}.

Recent benchmarks place greater emphasis on answer coverage and information integration. DeepSearchQA~\cite{Gupta2026DeepSearchQABT} requires agents to follow linked search steps, gather information from different sources, and produce complete answer lists. GISA supports several answer formats, including individual items, sets, lists, and tables. It also provides human search trajectories to support process analysis and model training~\cite{Zhu2026GISAAB}. DeepResearch Bench II~\cite{Li2026DeepResearchBI} extends evaluation to long-form research reports. It uses detailed scoring criteria derived from expert reports to assess information coverage, analysis, and presentation.

Research on search methods also examines how agents organize evidence and adjust later search actions. SearchOS~\cite{zhang2026searchos} maintains shared records of pending tasks, evidence, information coverage, and failed attempts. These records help coordinate multiple agents, fill information gaps, and reduce repeated unsuccessful searches. Further work uses search trajectories for supervised fine-tuning to connect retrieval decisions more closely with the reasoning process~\cite{Chen2026LearningTS}.

\subsection{Multimodal Search and Browsing}

BrowseComp~\cite{Wei2025BrowseCompAS} mainly evaluates text-based information seeking on the open web and provides a reference for assessing browsing capabilities. MMSearch~\cite{Jiang2024MMSearchBT} introduces image-text queries and evaluates query rewriting, result reranking, information summarization, and end-to-end answer generation. BrowseComp-VL~\cite{Geng2025WebWatcherBN} combines visual recognition with complex web retrieval, requiring agents to conduct multi-step searches based on entities in images. Later benchmarks place greater emphasis on the role of visual evidence during search. MMSearch-Plus~\cite{Tao2025MMSearchPlusBP} requires agents to extract fine-grained visual clues and verify related facts and sources through multiple rounds of image-text retrieval. VDR-Bench~\cite{Zeng2026VisionDeepResearchBR} focuses on reducing shortcuts based on textual clues and whole-image matching. It uses retrieval from image regions, entity verification, and multi-hop question construction to place greater demands on visual search. VisBrowse-Bench~\cite{Zhang2026VisBrowseBenchBV} emphasizes actively gathering visual information from webpages and evaluates spatial localization, attribute recognition, and reasoning across images.

Other work further expands search structures and process evaluation. InterLV-Search~\cite{Hou2026InterLVSearchBI} examines how visual and textual evidence guide later search steps. Its open-web tasks include single-chain and multi-branch structures that involve comparing multiple objects or search paths. MM-BrowseComp~\cite{Li2025MMBrowseCompAC} introduces multimodal evidence in both the input and the browsing process. It uses reasoning checklists to assess the completion of key steps. BrowseComp-V3~\cite{Zhang2026BrowseCompV3AV} covers alignment within image regions, integration across regions, and reasoning across images. It requires publicly searchable evidence and evaluates both final-answer accuracy and subgoal completion.

VisSearch-Bench further focuses on visual context management in search involving multiple images, multiple objects, and linked steps. Its tasks involve checking objects individually, combining results across a set, following visual clues, and distinguishing similar candidates. They also include using image structure together with external information to determine search targets. Table~\ref{tab:benchmark-comparison} compares the main search requirements covered by VisSearch-Bench and existing benchmarks based on their task designs.

\begin{figure}[!t]
\centering
\includegraphics[width=\textwidth]{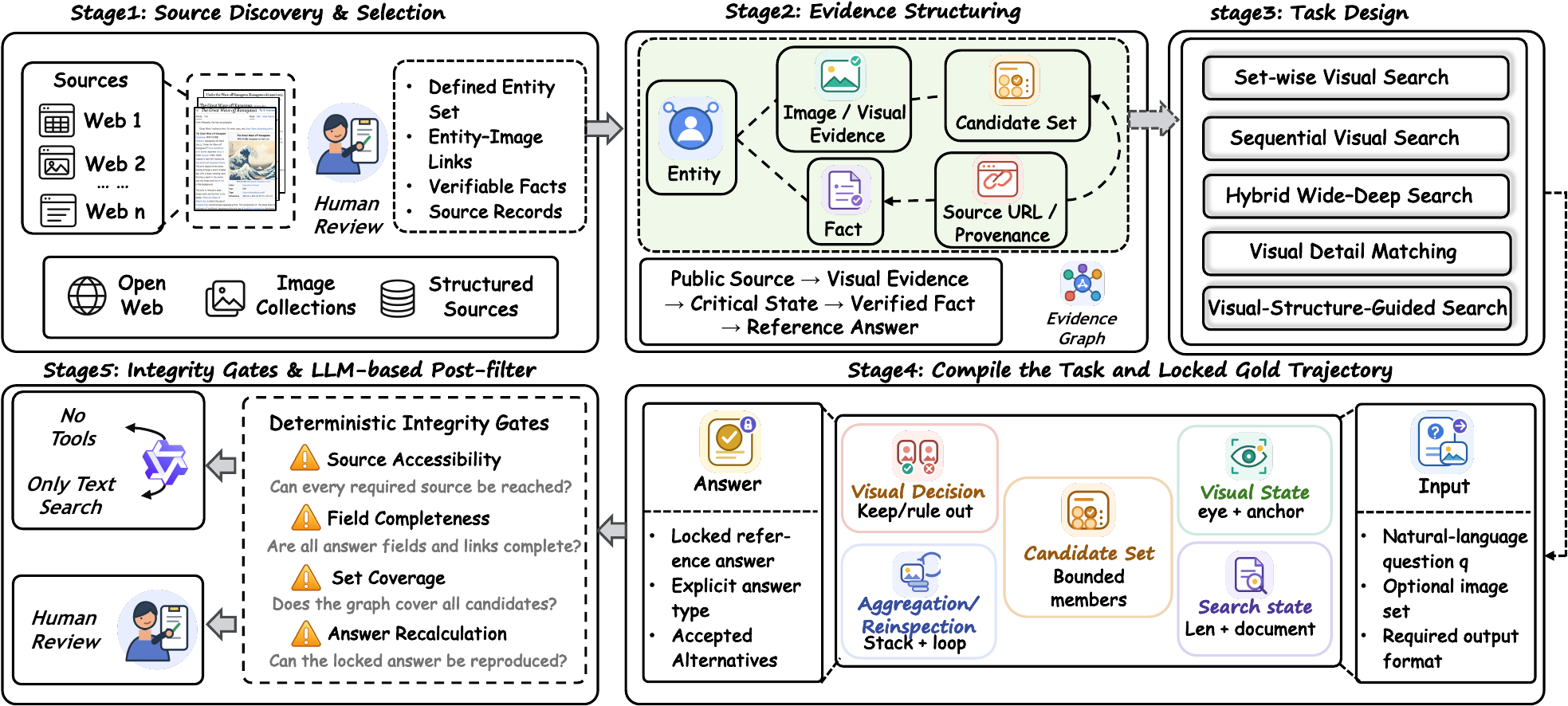}
\caption{Overview of the VisSearch-Bench construction pipeline. Human annotators curate public sources and organize them into traceable evidence graphs; structural constraints instantiate the five task families; each task is paired with a locked answer and gold trajectory; deterministic integrity gates, LLM-based post-filtering, and final human review verify the retained items.}
\label{fig:construction-pipeline}
\end{figure}

\section{VisSearch-Bench}


\subsection{Task Definition and Taxonomy}

We study search in open-web environments where visual context plays a central role. Each task provides a natural-language question \(q\), an optional initial image set \(\mathcal{I}_0\) that is empty for text-initiated tasks, and a target output format \(\sigma\). A multimodal agent interacts with the web through tools for text search, image search, reverse image search, webpage access, and image inspection. At step \(t\), the agent selects its next action \(a_{t+1}\) based on the interaction history

\begin{equation}
h_t=\bigl(q,\mathcal{I}_0,(a_1,o_1),\ldots,(a_t,o_t)\bigr)
\end{equation}

where \(o_t\) denotes the textual or visual observation returned by a tool. When the interaction ends, the agent produces a predicted answer \(\hat{y}\) in the required format \(\sigma\).

In VisSearch-Bench, we define \textbf{visual context} as the visual information involved in a task and its connections to object identities, external facts, and search targets. This information may come from input images or from images and webpages found during search. It includes object appearance and local features, as well as the order, positions, and connections of objects. \textbf{Visual context management} refers to organizing, connecting, and using this information to solve a task. This includes checking objects, comparing candidates, filtering results, and combining findings across steps. Based on these requirements, we organize VisSearch-Bench into five task categories. Each category emphasizes a main search requirement.

\begin{figure}[!t]
\centering
\includegraphics[width=\textwidth,height=0.30\textheight,keepaspectratio]{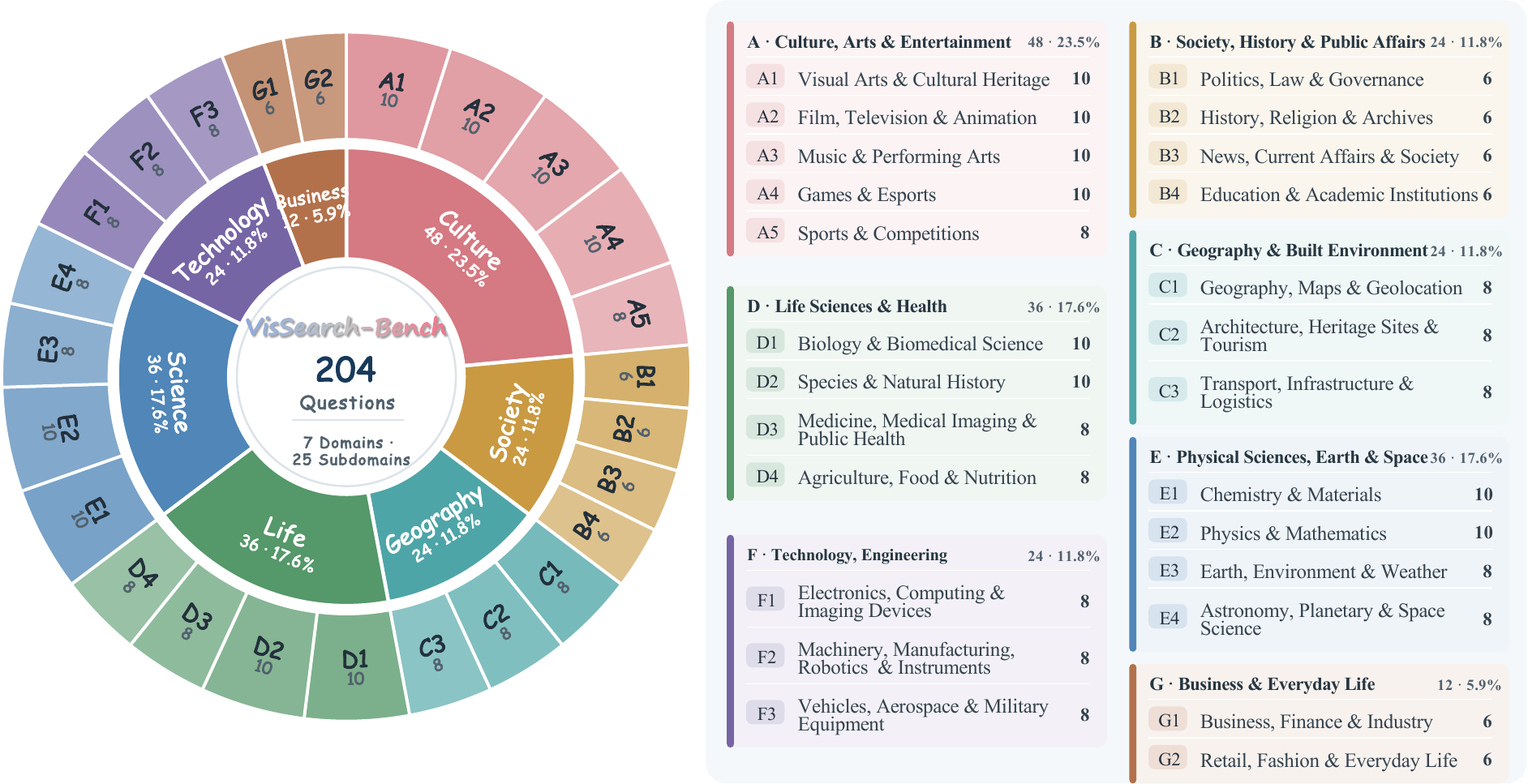}
\caption{Domain taxonomy and composition of VisSearch-Bench. The inner ring shows seven primary domains, and the outer ring shows 25 secondary domains. Arc area is proportional to the number of tasks in the 204-question benchmark. The right panel expands the full taxonomy and reports the task count within each category.}
\label{fig:domain-taxonomy}
\end{figure}

\begin{enumerate}
\renewcommand{\labelenumi}{\arabic{enumi})}
\setlength{\itemsep}{0.35em}
\setlength{\parsep}{0pt}
\setlength{\topsep}{0.3em}
\item \textbf{Set-wise Visual Search.} These tasks involve a clearly defined
set of objects. The set may come from a catalog specified in the question or
consist of objects shown in the input images. For each object, the agent
gathers and checks visual information. It then combines this information with
relevant facts to filter, compare, or summarize the results. The main challenge
is to cover all required objects and correctly link the information for each
one. Individual judgments must also remain consistent with the final summary.
\item \textbf{Sequential Visual Search.} These tasks involve multiple search
steps that depend on one another. An object, visual clue, or fact found in one
step guides later searches. For example, an agent may first identify an
artwork in an image, then find records about its artist, collection, or
exhibitions. Different steps may involve image understanding or text search.
The focus is on correctly connecting the information found across steps to
produce the required answer.
\item \textbf{Hybrid Wide--Deep Search.} These tasks combine sequential search
with search over a set of objects. Earlier findings can define the candidate
set to be checked. Filtering this set can then determine the target of the
next stage. For example, an agent may start from one work and identify its
award and year. It then examines images of works from the same award cycle
and selects those that meet the given conditions. Further searches gather
information about the selected works. The main challenge is to correctly
connect the search scope, individual judgments, and later targets across stages.
\item \begin{minipage}[t]{\linewidth}
\textbf{Competitive Visual Disambiguation.} These tasks focus on visual
targets that are easily confused with other objects. Candidates may look
similar overall but differ in local shapes, textures, markings, or individual
parts. The agent compares the input image with retrieved images, descriptions,
and source records to verify the target's identity. These tasks assess the
accuracy of visual detail matching and identity verification.
\end{minipage}
\item \begin{minipage}[t]{\linewidth}
\textbf{Visual-Structure-Guided Search.} These tasks use object order, spatial
positions, or connections in an image to determine what to search for, in what
order, or where to search next. The agent combines this structural information
with external search results to locate objects, check conditions, and summarize
findings. The main challenge is to correctly connect image structure, object
identities, and search results, then use these connections to complete the task.
\end{minipage}
\end{enumerate}

\subsection{Dataset Construction}

\begin{wraptable}{r}{0.52\textwidth}
\centering
\fontsize{9}{11}\selectfont
\setlength{\tabcolsep}{2.5pt}
\renewcommand{\arraystretch}{1.12}
\begin{tabularx}{\linewidth}{@{}>{\raggedright\arraybackslash}Xr@{}}
\toprule
\textbf{Statistic} & \textbf{Value} \\
\midrule
Total questions & 204 \\
Number of task families & 5 \\
Number of primary domains & 7 \\
\midrule
\textbf{Set-wise Visual Search} &
68 (33.3\%) \\
\hspace{0.7em}Reference-solution steps (avg.\ / max.) &
23.9 / 51 \\
\arrayrulecolor{vissearchrule}\specialrule{0.3pt}{1pt}{1pt}\arrayrulecolor{black}
\textbf{Sequential Visual Search} &
34 (16.7\%) \\
\hspace{0.7em}Reference-solution steps (avg.\ / max.) &
14.4 / 19 \\
\arrayrulecolor{vissearchrule}\specialrule{0.3pt}{1pt}{1pt}\arrayrulecolor{black}
\textbf{Hybrid Wide--Deep Search} &
34 (16.7\%) \\
\hspace{0.7em}Reference-solution steps (avg.\ / max.) &
23.8 / 61 \\
\arrayrulecolor{vissearchrule}\specialrule{0.3pt}{1pt}{1pt}\arrayrulecolor{black}
\textbf{Competitive Visual Disambiguation} &
34 (16.7\%) \\
\hspace{0.7em}Reference-solution steps (avg.\ / max.) &
3.9 / 5 \\
\arrayrulecolor{vissearchrule}\specialrule{0.3pt}{1pt}{1pt}\arrayrulecolor{black}
\textbf{Visual-Structure-Guided Search} &
34 (16.7\%) \\
\hspace{0.7em}Reference-solution steps (avg.\ / max.) &
15.8 / 19 \\
\midrule
Tasks with initial images & 187 (91.7\%) \\
Tasks without initial images & 17 (8.3\%) \\
Total reference-solution steps & 3{,}593 \\
Reference-solution steps per task (median / mean) & 17.0 / 17.6 \\
\bottomrule
\end{tabularx}
\captionsetup{font=footnotesize,skip=4pt}
\caption{Dataset composition and structural complexity.}
\label{tab:dataset-composition}
\end{wraptable}
Figure~\ref{fig:construction-pipeline} summarizes our human-led construction
process with program-assisted verification. The process follows three
principles. \textbf{Visual Indispensability} means that critical visual
evidence cannot be replaced by the question text, webpage titles, captions,
OCR, or model knowledge. \textbf{Visual Evidence Integration} means that tasks
require agents to connect visual observations with object identities and external
information. Agents use these connections to guide searches, compare candidates,
filter objects, or combine results across steps. \textbf{Public Solvability
and Auditability} means that each task uses publicly retrievable evidence and
has verifiable critical states and a clear reference answer.

Annotators first select public sources that provide a bounded entity set,
reliable entity-image links, and independently verifiable factual fields. We
fix these sources before question writing. The reference answer therefore does
not depend on model search results or subjective judgments. Annotators then
organize the entities, images, factual fields, and candidate sets into a
traceable evidence structure. They apply the structural constraints of one
task family from Section~3.1 and write a question that hides the intermediate
nodes and final answer. Automated checks verify source accessibility, field
completeness, set coverage, answer recomputation, and possible shortcuts. We
discard tasks that bypass critical visual evidence or allow a shorter
unintended path. Human reviewers inspect every retained task.

\subsection{Evaluation Protocol and Metrics}

Our primary metric is \textbf{Accuracy (Acc.)}. For task \(i\), let \(y_i\),
\(\hat{y}_i\), and \(\sigma_i\) denote the reference answer, the predicted
answer, and the required answer format. We define
\begin{gather}
a_i = \mathbb{I}\!\left[
\mathcal{E}_{\sigma_i}(\hat{y}_i,y_i)=1\right],
\label{eq:answer-correctness}\\
\mathrm{Acc} = \frac{1}{N}\sum_{i=1}^{N}a_i.
\label{eq:accuracy}
\end{gather}
where \(\mathcal{E}_{\sigma_i}\) compares parsed answers under predefined,
task-specific rules. It normalizes entity aliases. For unordered sets, it
removes duplicates and ignores order. For ordered sequences, it preserves
order. A task is correct only if the prediction contains all and only the
required content.

\begin{table}[!t]
\centering
\fontsize{8.5}{10}\selectfont
\setlength{\tabcolsep}{2.35pt}
\renewcommand{\arraystretch}{1.12}
\makebox[\textwidth][c]{%
\resizebox{1.06\textwidth}{!}{%
\begin{tabular}{@{}l>{\columncolor[HTML]{F3EFFA}}c*{7}{c}@{\hspace{4pt}}>{\columncolor[HTML]{F3EFFA}}c*{7}{c}@{}}
\toprule
\textbf{Model / Backbone}
& \multicolumn{8}{c}{\textbf{Accuracy (Acc., \%)}}
& \multicolumn{8}{c}{\textbf{Process Score (\%)}} \\
\cmidrule(lr){2-9}\cmidrule(lr){10-17}
& \textbf{Avg.} & \textbf{Cult.} & \textbf{Soc.} & \textbf{Geo.}
& \textbf{Life} & \textbf{Sci.} & \textbf{Tech.} & \textbf{Bus.}
& \textbf{Avg.} & \textbf{Cult.} & \textbf{Soc.} & \textbf{Geo.}
& \textbf{Life} & \textbf{Sci.} & \textbf{Tech.} & \textbf{Bus.} \\
\midrule
\rowcolor{vissearchgroup}
\multicolumn{17}{c}{\textbf{Open-source Models}} \\
Qwen3.6-35B-A3B
& 0.98 & 0.00 & 0.00 & 0.00 & 5.56 & 0.00 & 0.00 & 0.00
& 9.55 & 4.19 & 10.50 & 9.94 & 17.48 & 12.17 & 6.46 & 2.92 \\
Qwen3.6-27B
& 0.00 & 0.00 & 0.00 & 0.00 & 0.00 & 0.00 & 0.00 & 0.00
& 5.53 & 3.22 & 9.79 & 6.02 & 8.46 & 4.15 & 2.29 & 7.02 \\
Qwen3.5-397B-A17B
& 1.96 & 0.00 & 0.00 & 0.00 & 11.11 & 0.00 & 0.00 & 0.00
& 12.72 & 7.00 & 13.75 & 11.05 & 21.70 & 11.93 & 13.46 & 10.78 \\
GLM-4.6V
& 0.49 & 0.00 & 0.00 & 0.00 & 2.78 & 0.00 & 0.00 & 0.00
& 9.24 & 4.14 & 10.53 & 8.93 & 15.82 & 7.14 & 12.26 & 8.24 \\
MiniMax-M3
& 2.45 & 2.08 & 0.00 & 4.17 & 8.33 & 0.00 & 0.00 & 0.00
& 9.71 & 6.69 & 9.46 & 5.59 & 18.03 & 9.21 & 10.98 & 4.46 \\
Qwen3-VL-235B-A22B-Thinking
& 1.47 & 0.00 & 0.00 & 4.17 & 2.78 & 2.78 & 0.00 & 0.00
& 11.37 & 6.60 & 14.41 & 10.13 & 16.28 & 14.00 & 11.64 & 3.61 \\
Qwen3-VL-8B-Instruct
& 0.49 & 0.00 & 0.00 & 0.00 & 2.78 & 0.00 & 0.00 & 0.00
& 5.83 & 1.74 & 13.62 & 1.82 & 13.84 & 1.58 & 6.13 & 2.78 \\
Qwen3-VL-4B-Instruct
& 0.00 & 0.00 & 0.00 & 0.00 & 0.00 & 0.00 & 0.00 & 0.00
& 5.17 & 0.54 & 10.09 & 4.23 & 9.60 & 4.57 & 5.95 & 2.78 \\
GLM-5.3-Flash
& 3.92 & 6.25 & 0.00 & 4.17 & 8.33 & 0.00 & 4.17 & 0.00
& 8.66 & 8.75 & 2.50 & 8.13 & 17.18 & 5.96 & 9.03 & 3.51 \\
Qwen3.8-Flash-Next
& 3.43 & 6.25 & 0.00 & 4.17 & 5.56 & 0.00 & 0.00 & \textbf{8.33}
& 11.82 & 8.06 & 8.36 & 16.02 & 19.06 & 9.07 & 11.91 & 11.75 \\
Qwen3.8-27B
& 2.45 & 0.00 & 0.00 & 4.17 & 11.11 & 0.00 & 0.00 & 0.00
& 6.35 & 0.35 & 3.06 & 8.46 & 14.82 & 6.72 & 5.11 & 8.70 \\
DeepSeek-V4.1-Flash
& 1.96 & 0.00 & 4.17 & 0.00 & 5.56 & 0.00 & 0.00 & \textbf{8.33}
& 12.88 & 6.93 & 13.42 & 8.22 & 19.33 & 13.07 & 19.00 & 12.84 \\
Kimi-K3
& 5.88 & 8.33 & 4.17 & 4.17 & 11.11 & 2.78 & 4.17 & 0.00
& 10.24 & 7.73 & 9.12 & 8.07 & 19.26 & 9.31 & 11.58 & 0.00 \\
\midrule
\rowcolor{vissearchgroup}
\multicolumn{17}{c}{\textbf{Closed-source Models}} \\
Gemini 3.5 Flash
& 6.37 & 6.25 & 8.33 & 8.33 & 8.33 & 2.78 & 4.17 & \textbf{8.33}
& 22.28 & 17.42 & 23.11 & 29.11 & 27.59 & 19.29 & 24.39 & 15.22 \\
Claude Opus 4.8
& 4.90 & 4.17 & 8.33 & 4.17 & 5.56 & 5.56 & 0.00 & \textbf{8.33}
& 16.13 & 12.40 & 17.51 & 11.12 & 18.27 & 18.21 & 17.51 & 22.97 \\
GPT-5.5
& 6.37 & 8.33 & 4.17 & 4.17 & 11.11 & 5.56 & 4.17 & 0.00
& 22.61 & 19.19 & 18.83 & 30.08 & 25.33 & 18.46 & 32.95 & 12.53 \\
\midrule
\rowcolor{vissearchgroup}
\multicolumn{17}{c}{\textbf{Open-source Agents}} \\
\rowcolor{vissearchsubgroup}
\multicolumn{17}{c}{\itshape ReAct} \\
Qwen3.6-35B-A3B
& 7.35 & 4.17 & 8.33 & 8.33 & 5.56 & 8.33 & 16.67 & 0.00
& 23.25 & 14.44 & 22.74 & 27.27 & 25.56 & 24.91 & 29.63 & 26.77 \\
GPT-5.5
& 25.49 & 20.83 & 20.83 & \textbf{33.33} & 33.33 & \textbf{25.00} & \textbf{29.17} & \textbf{8.33}
& \textbf{49.52} & \textbf{46.70} & 43.42 & \textbf{60.37} & \textbf{52.76} & \textbf{48.02} & \textbf{59.05} & 27.02 \\
Qwen3.5-397B-A17B
& 15.69 & 16.67 & 16.67 & 8.33 & 16.67 & 16.67 & 25.00 & 0.00
& 32.43 & 26.14 & 30.82 & 26.67 & 33.47 & 42.68 & 41.19 & 20.89 \\
GLM-5.3-Flash
& 20.10 & 20.83 & 16.67 & 20.83 & 36.11 & 13.89 & 16.67 & 0.00
& 22.29 & 21.70 & 19.58 & 20.83 & 36.11 & 19.04 & 22.92 & 0.00 \\
Kimi-K3
& \textbf{26.47} & \textbf{25.00} & \textbf{33.33} & \textbf{33.33} & 36.11 & 22.22 & 20.83 & 0.00
& 38.57 & 39.50 & \textbf{45.00} & 40.00 & 42.75 & 34.58 & 37.06 & 21.60 \\
DeepSeek-V4.1-Flash
& 22.55 & 14.58 & 20.83 & 29.17 & \textbf{38.89} & 19.44 & 25.00 & 0.00
& 39.47 & 38.03 & 33.13 & 38.44 & 50.12 & 37.50 & 44.83 & 23.18 \\
Qwen3.8-27B
& 21.08 & 16.67 & 25.00 & 20.83 & 36.11 & 13.89 & 20.83 & \textbf{8.33}
& 33.23 & 26.28 & 34.25 & 32.12 & 41.48 & 32.70 & 37.05 & 30.35 \\
\rowcolor{vissearchsubgroup}
\multicolumn{17}{c}{\itshape DeerFlow} \\
Qwen3.6-35B-A3B
& 10.78 & 10.42 & 12.50 & 4.17 & 11.11 & 13.89 & 16.67 & 0.00
& 24.26 & 17.89 & 26.45 & 17.21 & 29.03 & 33.15 & 27.77 & 11.49 \\
GPT-5.5
& 20.10 & 16.67 & 29.17 & 12.50 & 27.78 & 16.67 & 25.00 & \textbf{8.33}
& 38.30 & 35.34 & 43.86 & 35.39 & 34.78 & 35.33 & 51.41 & \textbf{38.13} \\
Qwen3.5-397B-A17B
& 10.29 & 8.33 & 12.50 & 0.00 & 19.44 & 11.11 & 12.50 & 0.00
& 28.37 & 19.54 & 32.44 & 22.59 & 33.13 & 40.88 & 29.44 & 13.11 \\
\bottomrule
\end{tabular}
}
}
\caption{Main results on VisSearch-Bench. Avg.\ denotes the micro-average over all benchmark tasks. Cult., Soc., Geo., Life, Sci., Tech., and Bus.\ denote Culture, Society, Geography, Life Sciences, Physical Sciences, Technology, and Business, respectively.}
\label{tab:main-results}
\end{table}

Final-answer accuracy cannot show whether an error comes from the wrong search
space, a missed visual candidate, or incorrect evidence
aggregation~\cite{wu2026silent}. We therefore also report
\textbf{Process Score (PS)}. It measures the correctness and completeness of
individual results and answers to subquestions, using the scoring rules defined
for each task. Each task
\(i\) has \(m_i\) labeled critical states, which all systems report before the
final answer. Let \([m_i]=\{1,\ldots,m_i\}\) index these states. We define the
set of correctly recovered states as
\begin{gather}
\mathcal{R}_i =
\left\{j\in[m_i]\;\middle|\;
\hat{g}_{ij}\equiv_{\tau_{ij}}g_{ij}\right\},
\label{eq:recovered-states}\\
\mathrm{PS} = \frac{1}{N}\sum_{i=1}^{N}
\frac{|\mathcal{R}_i|}{m_i}.
\label{eq:process-score}
\end{gather}
Here, \(\hat{g}_{ij}\equiv_{\tau_{ij}}g_{ij}\) means that predicted state \(j\) matches its reference under the corresponding rule. States are aligned by field labels, and different names for the same entity are treated as equal. Each match counts as one correct state, so \(|\mathcal{R}_i|/m_i\) is the fraction of required states recovered for task \(i\). We average the process scores across all tasks to obtain the overall Process Score for the benchmark. Different search paths receive the same score if they recover the same critical states. Direct uses no tools but reports the same structured states and is scored in the same way. All tables report both metrics as percentages.

\begin{table}[!t]
\centering
\fontsize{8.5}{10}\selectfont
\setlength{\tabcolsep}{2.35pt}
\renewcommand{\arraystretch}{1.12}
\makebox[\textwidth][c]{%
\resizebox{1.06\textwidth}{!}{%
\begin{tabular}{@{}l*{2}{>{\columncolor[HTML]{F3EFFA}}c}*{5}{@{\hspace{4pt}}cc}@{}}
\toprule
\textbf{Model / Backbone}
& \multicolumn{2}{>{\columncolor[HTML]{F3EFFA}}c}{\textbf{\shortstack{Avg.\\(204)}}}
& \multicolumn{2}{c}{\textbf{\shortstack{Set-wise\\Visual Search\\(68)}}}
& \multicolumn{2}{c}{\textbf{\shortstack{Sequential\\Visual Search\\(34)}}}
& \multicolumn{2}{c}{\textbf{\shortstack{Hybrid Wide--Deep\\Search\\(34)}}}
& \multicolumn{2}{c}{\textbf{\shortstack{Competitive Visual\\Disambiguation\\(34)}}}
& \multicolumn{2}{c}{\textbf{\shortstack{Visual-Structure-\\Guided Search\\(34)}}} \\
\cmidrule(lr){2-3}\cmidrule(lr){4-5}\cmidrule(lr){6-7}\cmidrule(lr){8-9}\cmidrule(lr){10-11}\cmidrule(lr){12-13}
& \textbf{Acc.} & \textbf{\shortstack{Process\\Score}} & \textbf{Acc.} & \textbf{\shortstack{Process\\Score}} & \textbf{Acc.} & \textbf{\shortstack{Process\\Score}} & \textbf{Acc.} & \textbf{\shortstack{Process\\Score}} & \textbf{Acc.} & \textbf{\shortstack{Process\\Score}} & \textbf{Acc.} & \textbf{\shortstack{Process\\Score}} \\
\midrule
\rowcolor{vissearchgroup}
\multicolumn{13}{c}{\textbf{Open-source Models}} \\
Qwen3.6-35B-A3B
& 0.98 & 9.55 & 0.00 & 10.79 & 0.00 & 6.40 & 0.00 & 5.42 & 0.00 & 0.00 & 5.88 & 23.94 \\
Qwen3.6-27B
& 0.00 & 5.53 & 0.00 & 7.94 & 0.00 & 5.44 & 0.00 & 1.53 & 0.00 & 0.00 & 0.00 & 10.31 \\
Qwen3.5-397B-A17B
& 1.96 & 12.72 & 0.00 & 13.77 & 0.00 & 10.48 & 0.00 & 11.12 & 0.00 & 0.00 & 11.76 & 27.17 \\
GLM-4.6V
& 0.49 & 9.24 & 0.00 & 6.94 & 0.00 & 5.29 & 0.00 & 10.74 & 0.00 & 0.00 & 2.94 & 25.53 \\
MiniMax-M3
& 2.45 & 9.71 & 0.00 & 7.18 & 0.00 & 7.38 & 2.94 & 4.89 & 0.00 & 0.00 & 11.76 & 31.61 \\
Qwen3-VL-235B-A22B-Thinking
& 1.47 & 11.37 & 0.00 & 13.21 & 0.00 & 7.28 & 0.00 & 7.22 & 0.00 & 0.00 & 8.82 & 27.27 \\
Qwen3-VL-8B-Instruct
& 0.49 & 5.83 & 0.00 & 7.15 & 0.00 & 2.85 & 0.00 & 4.07 & 0.00 & 0.00 & 2.94 & 13.77 \\
Qwen3-VL-4B-Instruct
& 0.00 & 5.17 & 0.00 & 6.69 & 0.00 & 2.62 & 0.00 & 3.35 & 0.00 & 0.00 & 0.00 & 11.69 \\
GLM-5.3-Flash
& 3.92 & 8.66 & 0.00 & 4.22 & 0.00 & 4.26 & 8.82 & 8.82 & 0.00 & 0.00 & 14.71 & 30.45 \\
Qwen3.8-Flash-Next
& 3.43 & 11.82 & 0.00 & 6.24 & 0.00 & 9.56 & 8.82 & 14.94 & 0.00 & 0.00 & 11.76 & 33.96 \\
Qwen3.8-27B
& 2.45 & 6.35 & 0.00 & 4.33 & 0.00 & 1.03 & 0.00 & 0.41 & 0.00 & 0.00 & 14.71 & 28.02 \\
DeepSeek-V4.1-Flash
& 1.96 & 12.88 & 0.00 & 11.69 & 0.00 & 11.54 & 2.94 & 13.72 & 0.00 & 0.00 & 8.82 & 28.66 \\
Kimi-K3
& 5.88 & 10.24 & 2.94 & 6.69 & 2.94 & 11.47 & 8.82 & 10.82 & 0.00 & 0.00 & 17.65 & 25.79 \\
\midrule
\rowcolor{vissearchgroup}
\multicolumn{13}{c}{\textbf{Closed-source Models}} \\
Gemini 3.5 Flash
& 6.37 & 22.28 & 0.00 & 21.40 & 2.94 & 22.35 & 11.76 & 28.08 & 0.00 & 0.00 & 23.53 & 40.44 \\
Claude Opus 4.8
& 4.90 & 16.13 & 0.00 & 12.28 & 5.88 & 15.14 & 11.76 & 20.02 & 0.00 & 0.00 & 11.76 & 37.09 \\
GPT-5.5
& 6.37 & 22.61 & 1.47 & 21.23 & 2.94 & 21.36 & 8.82 & 28.36 & 0.00 & 0.00 & 23.53 & 43.47 \\
\midrule
\rowcolor{vissearchgroup}
\multicolumn{13}{c}{\textbf{Open-source Agents}} \\
\rowcolor{vissearchsubgroup}
\multicolumn{13}{c}{\itshape ReAct} \\
Qwen3.6-35B-A3B
& 7.35 & 23.25 & 2.94 & 19.81 & 2.94 & 16.47 & 5.88 & 23.40 & 2.94 & 2.94 & 26.47 & 57.05 \\
GPT-5.5
& 25.49 & \textbf{49.52} & 8.82 & \textbf{44.10} & 23.53 & 51.73 & \textbf{23.53} & \textbf{48.36} & 38.24 & 38.24 & \textbf{50.00} & \textbf{70.61} \\
Qwen3.5-397B-A17B
& 15.69 & 32.43 & 0.00 & 20.14 & 11.76 & 41.60 & 14.71 & 27.21 & 32.35 & 29.41 & 35.29 & 56.05 \\
GLM-5.3-Flash
& 20.10 & 22.29 & 10.29 & 11.58 & 8.82 & 15.29 & 8.82 & 8.82 & 35.29 & 35.29 & 47.06 & 51.16 \\
Kimi-K3
& \textbf{26.47} & 38.57 & \textbf{11.76} & 24.00 & \textbf{32.35} & \textbf{56.34} & 14.71 & 27.74 & 41.18 & \textbf{44.12} & 47.06 & 55.22 \\
DeepSeek-V4.1-Flash
& 22.55 & 39.47 & 5.88 & 28.28 & 26.47 & 52.52 & 11.76 & 20.97 & \textbf{44.12} & \textbf{44.12} & 41.18 & 62.63 \\
Qwen3.8-27B
& 21.08 & 33.23 & 7.35 & 18.78 & 23.53 & 46.91 & 8.82 & 20.17 & \textbf{44.12} & \textbf{44.12} & 35.29 & 50.61 \\
\rowcolor{vissearchsubgroup}
\multicolumn{13}{c}{\itshape DeerFlow} \\
Qwen3.6-35B-A3B
& 10.78 & 24.26 & 1.47 & 19.08 & 8.82 & 20.29 & 5.88 & 17.55 & 35.29 & 35.29 & 11.76 & 34.27 \\
GPT-5.5
& 20.10 & 38.30 & 4.41 & 31.35 & 17.65 & 40.76 & 20.59 & 39.75 & 26.47 & 26.47 & 47.06 & 60.14 \\
Qwen3.5-397B-A17B
& 10.29 & 28.37 & 0.00 & 17.59 & 8.82 & 37.65 & 8.82 & 23.80 & 29.41 & 29.41 & 14.71 & 44.18 \\
\bottomrule
\end{tabular}
}
}
\caption{Accuracy and Process Score (\%) by task category on VisSearch-Bench. Numbers in parentheses indicate task counts. Avg.\ denotes the micro-average over all 204 tasks. Bold values mark the highest score in each column, including ties.}
\label{tab:task-family-accuracy}
\end{table}

\subsection{Dataset Statistics}

VisSearch-Bench contains 204 tasks organized along two dimensions. The first dimension is search structure. It includes five task categories: Set-wise Visual Search, Sequential Visual Search, Hybrid Wide--Deep Search, Competitive Visual Disambiguation, and Visual-Structure-Guided Search. Each category emphasizes a different search requirement. Set-wise Visual Search contains 68 tasks, or 33.3\% of the benchmark. Each of the other four families contains 34 tasks, or 16.7\%. The second is content domain, with 7 primary domains and 25 secondary domains that measure coverage across knowledge areas and real-world settings.

The seven primary domains are Culture (23.5\%), Life Sciences (17.6\%), Physical Sciences (17.6\%), Society (11.8\%), Geography (11.8\%), Technology (11.8\%), and Business (5.9\%). Each task is assigned one primary and one secondary domain based on its entities and required knowledge. Figure~\ref{fig:domain-taxonomy} shows the taxonomy for all 25 secondary domains.

\begin{table}[!t]
\centering
\fontsize{9}{11}\selectfont
\setlength{\tabcolsep}{3pt}
\renewcommand{\arraystretch}{1.22}
\makebox[\textwidth][c]{%
\resizebox{1.06\textwidth}{!}{%
\begin{tabular}{@{}l*{12}{c}@{}}
\toprule
\textbf{Model / Backbone}
& \multicolumn{2}{c}{\textbf{Direct Answer}}
& \multicolumn{2}{c}{\textbf{Text}}
& \multicolumn{2}{c}{\textbf{Caption}}
& \multicolumn{2}{c}{\textbf{Full}}
& \multicolumn{2}{c}{\textbf{Full $-$ Text}}
& \multicolumn{2}{c}{\textbf{Full $-$ Caption}} \\
\cmidrule(lr){2-3}\cmidrule(lr){4-5}\cmidrule(lr){6-7}\cmidrule(lr){8-9}\cmidrule(lr){10-11}\cmidrule(lr){12-13}
& \textbf{Acc.} & \textbf{\shortstack{Process\\Score}} & \textbf{Acc.} & \textbf{\shortstack{Process\\Score}} & \textbf{Acc.} & \textbf{\shortstack{Process\\Score}} & \textbf{Acc.} & \textbf{\shortstack{Process\\Score}} & \textbf{Acc.} & \textbf{\shortstack{Process\\Score}} & \textbf{Acc.} & \textbf{\shortstack{Process\\Score}} \\
\midrule
\rowcolor{vissearchgroup}
\multicolumn{13}{c}{\textbf{\itshape ReAct}} \\
Qwen3.6-35B-A3B
& 0.98 & 9.55 & 5.88 & 17.74 & 6.37 & 19.21 & \textbf{7.35} & \textbf{23.25} & +1.47 & +5.51 & +0.98 & +4.04 \\
Qwen3.5-397B-A17B
& 1.96 & 12.72 & 8.33 & 19.84 & 14.71 & 25.42 & \textbf{15.69} & \textbf{32.43} & +7.35 & +12.59 & +0.98 & +7.00 \\
DeepSeek-V4.1-Flash
& 1.96 & 12.88 & 12.25 & 23.14 & 19.12 & 36.33 & \textbf{22.55} & \textbf{39.47} & +10.29 & +16.33 & +3.43 & +3.14 \\
\bottomrule
\end{tabular}
}
}
\caption{Effect of visual access on VisSearch-Bench. Accuracy and Process Score are reported as percentages over all 204 tasks. Direct Answer uses no external tools; Text, Caption, and Full use ReAct. Full $-$ Text and Full $-$ Caption report gains in percentage points, computed before rounding.}
\label{tab:visual-indispensability}
\end{table}

\section{Experiments}

\subsection{Experimental Setup}

\paragraph{Models and agent frameworks.}
We evaluate 16 representative multimodal models. The open-weight models are
GLM-5.3-Flash, Kimi-K3~\cite{Bai2026KimiKO}, DeepSeek-V4.1-Flash~\cite{Xu2026DeepSeekV41FlashPT}, Qwen3.8-27B, Qwen3.8-Flash,
Qwen3.6-35B-A3B~\cite{qwen3.6-35b-a3b}, Qwen3.6-27B~\cite{qwen3.6-27b}, Qwen3.5-397B-A17B~\cite{qwen3.5}, GLM-4.6V~\cite{Hong2025GLM45VAG}, MiniMax-M3~\cite{Lai2026MiniMaxSA}, Qwen3-VL-235B-A22B-Thinking, Qwen3-VL-8B-Instruct, and
Qwen3-VL-4B-Instruct~\cite{Yang2025Qwen3TR,Bai2025Qwen3VLTR}. The proprietary models are GPT-5.5, Gemini 3.5 Flash,
and Claude Opus 4.8. In the Direct Answer setting, a model receives only the
question and its initial images. We also pair Qwen3.6-35B-A3B,
Qwen3.5-397B-A17B, GPT-5.5, GLM-5.3-Flash, Kimi-K3, DeepSeek-V4.1-Flash,
and Qwen3.8-27B with ReAct~\cite{Yao2022ReActSR}.
This comparison measures how much each search framework adds to the same
backbone.

\paragraph{Tools and inference limits.}
All ReAct agents have access to five evidence tools: web search, webpage
reading, text-based image search, reverse image search, and image cropping.
Each web, image, or reverse-image search returns up to five results.
Retrieved webpage text is truncated to 12,000 characters. Images shown to
the model have a longest side of at most 1,536 pixels. The image budget is
20 per task, shared by initial images, retrieved images, and crops. Crops
can be enlarged by up to four times. Each task allows up to 30 ReAct search
rounds and 60 evidence-tool calls, with at most four calls per round. If
either limit is reached, one additional model call is used to submit the
final answer from the available evidence.
\paragraph{Inference and evaluation.}
We evaluate each system once per task from a fresh context and report its first
complete response as Pass@1. The question, initial images, and required
state-and-answer schema are identical across settings; framework-specific
instructions only define the corresponding tool-use protocol. Forward web and
image search use Serper, whereas reverse-image search uses SerpApi.
We use the metrics defined in Section~3.3. Accuracy (Acc.) measures whether
the final answer is correct. Process Score measures the correctness and
completeness of individual results and answers to subquestions in the final
response, using task-specific scoring rules.

\subsection{Main Results}

Table~\ref{tab:main-results} reports the Pass@1 results on VisSearch-Bench.
In the Direct Answer setting, no model exceeds 6.37\% accuracy. With
search, ReAct with Kimi-K3 achieves the highest accuracy at 26.47\%.
ReAct with GPT-5.5 reaches 25.49\% accuracy and the highest Process Score
at 49.52\%. The best model therefore differs across the two metrics.
Success on complete tasks does not fully align with performance on
individual items and subquestions. Even the system with the highest
accuracy fully solves only about one quarter of the tasks. There is still
much room to improve how models link information across images, verify
objects, and combine results across search steps.

In the available paired experiments, both ReAct and DeerFlow improve
accuracy and Process Score over Direct Answer. For example,
Qwen3.5-397B-A17B achieves 1.96\% accuracy without tools. ReAct and
DeerFlow raise its accuracy to 15.69\% and 10.29\%, respectively. Its
Process Score also increases from 12.72\% to 32.43\% with ReAct and
28.37\% with DeerFlow. Thus, search improves both complete answers and
the correctness and completeness of individual results and subquestion
answers.

The effect of the framework also varies across models. For GPT-5.5 and
Qwen3.5-397B-A17B, ReAct outperforms DeerFlow on both metrics. For
Qwen3.6-35B-A3B, however, DeerFlow performs better. It achieves 10.78\%
accuracy and a Process Score of 24.26\%, compared with 7.35\% and 23.25\%
for ReAct. No single framework performs best for every model tested with
both frameworks.

\subsection{Ablation Studies}

\FloatBarrier
\begin{wraptable}{r}{0.52\textwidth}
\centering
\fontsize{8.5}{10.5}\selectfont
\setlength{\tabcolsep}{2.5pt}
\renewcommand{\arraystretch}{1.12}
\begin{tabularx}{\linewidth}{@{}l*{4}{>{\centering\arraybackslash}X}@{}}
\toprule
\textbf{Setting}
& \textbf{\shortstack{Full wrong\\$\rightarrow$ correct}}
& \textbf{\shortstack{Full correct\\$\rightarrow$ wrong}}
& \textbf{\shortstack{Correct\\answers}}
& \textbf{\shortstack{Acc.\\(\%)}} \\
\midrule
\rowcolor{vissearchgroup}
\multicolumn{5}{c}{\textbf{Qwen3.6-35B-A3B} \enspace (Full: 15, 7.35\%)} \\
Caption & 6 & \textbf{8} & 13 & 6.37 \\
Text & 6 & \textbf{9} & 12 & 5.88 \\
\midrule
\rowcolor{vissearchgroup}
\multicolumn{5}{c}{\textbf{Qwen3.5-397B-A17B} \enspace (Full: 32, 15.69\%)} \\
Caption & 8 & \textbf{10} & 30 & 14.71 \\
Text & 5 & \textbf{20} & 17 & 8.33 \\
\midrule
\rowcolor{vissearchgroup}
\multicolumn{5}{c}{\textbf{DeepSeek-V4.1-Flash} \enspace (Full: 46, 22.55\%)} \\
Caption & 13 & \textbf{20} & 39 & 19.12 \\
Text & 5 & \textbf{26} & 25 & 12.25 \\
\bottomrule
\end{tabularx}
\captionsetup{font=footnotesize,skip=4pt}
\caption{ReAct answer changes from Full to each setting on 204 tasks. Group headers give the number of correct answers and accuracy under Full.}
\label{tab:answer-changes}
\end{wraptable}

Table~\ref{tab:task-family-accuracy} reports accuracy and Process Score
across the five task categories. Accuracy on Set-wise Visual Search
remains low. ReAct with Kimi-K3 achieves the highest accuracy at 11.76\%.
In this category, ReAct with GPT-5.5 reaches a Process Score of 44.10\%,
but its accuracy is only 8.82\%. These results show a clear gap between
obtaining partially correct results and completing the entire set-based
task.

\begin{figure}[!t]
\centering
\includegraphics[width=\textwidth]{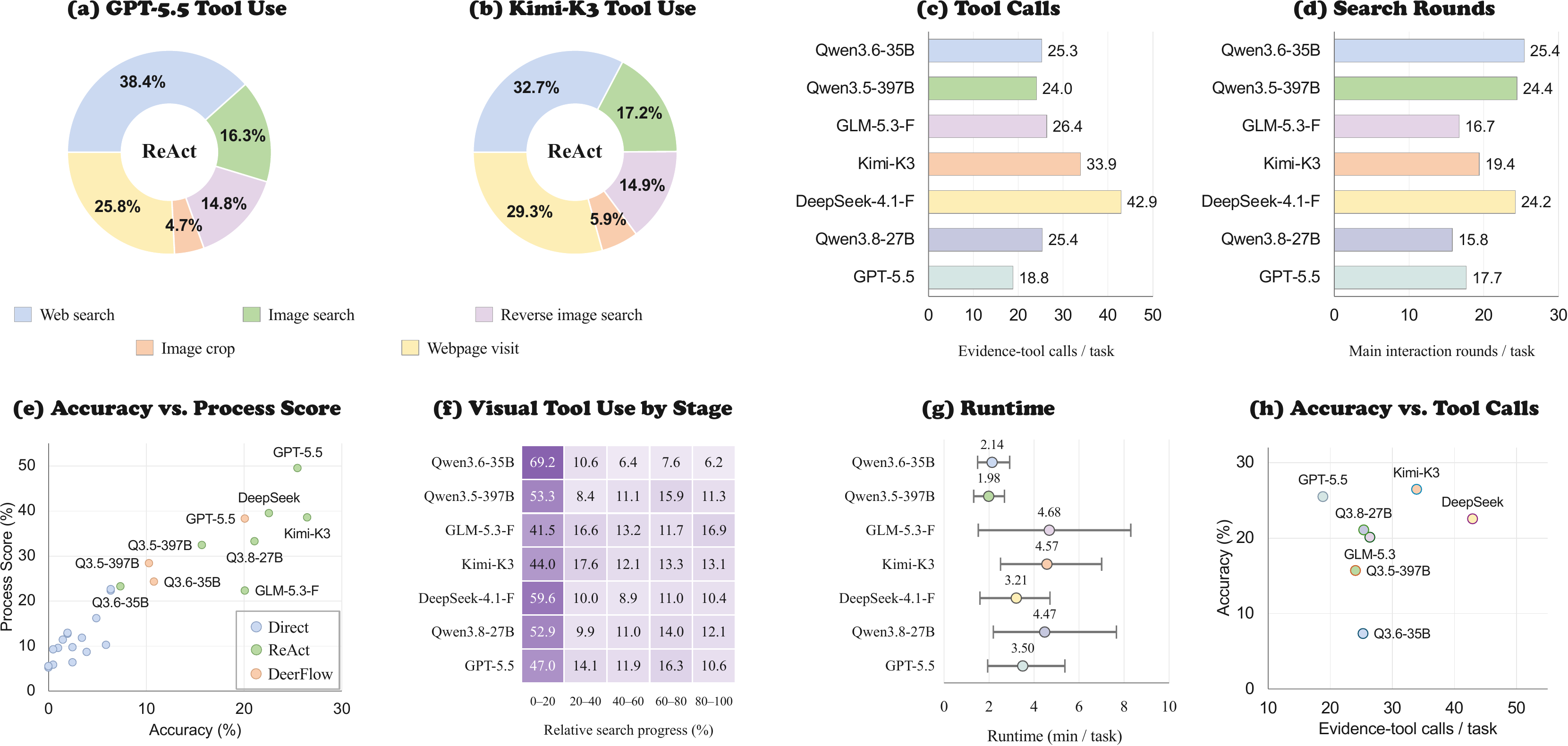}
\caption{Tool use, efficiency, and performance on VisSearch-Bench.}
\label{fig:tool-use-efficiency}
\end{figure}

Different models also show different strengths across the other
categories. ReAct with Kimi-K3 leads on Sequential Visual Search, with
32.35\% accuracy and a Process Score of 56.34\%. ReAct with GPT-5.5 leads
on both metrics in Hybrid Wide--Deep Search and Visual-Structure-Guided
Search. Its accuracy and Process Score are 23.53\% and 48.36\% for Hybrid
Wide--Deep Search. They reach 50.00\% and 70.61\% for
Visual-Structure-Guided Search. On Competitive Visual Disambiguation,
all Direct Answer models score 0.00\% accuracy. ReAct with
DeepSeek-V4.1-Flash and Qwen3.8-27B both reach 44.12\%. Search therefore
improves performance across different visual tasks, while the strengths
of individual models vary.


To study the role of different information sources, we compare four
settings. \emph{Direct Answer} receives only the question and its initial
images, without external tools. The other three settings use ReAct and
retain the same initial images. \emph{Text} provides web search and
webpage reading. \emph{Caption} also provides image search and reverse
image search. It uses separate calls to the same vision model to generate
general descriptions of retrieved images. This retains visual information
extraction but adds computation. \emph{Full} allows the agent to inspect
retrieved images directly and use image cropping.
Table~\ref{tab:visual-indispensability} shows that Full achieves the
highest accuracy and Process Score for all three models.

Table~\ref{tab:answer-changes} shows how gains and losses on different
tasks partly offset each other. Switching from Full to Caption loses
8 correct answers for Qwen3.6-35B-A3B, 10 for Qwen3.5-397B-A17B, and
20 for DeepSeek-V4.1-Flash. Caption also correctly answers 6, 8, and
13 tasks that Full fails, respectively. For tasks with demanding visual
context, the additional description step may reduce the burden of
organizing image information. This may help the agent solve some tasks
that fail under Full. However, general descriptions may omit details
needed for cross-image comparisons or other visual judgments. Across all
three models, Caption loses more correct answers than it gains. For
example, Qwen3.5-397B-A17B achieves 14.71\% accuracy with Caption, below
the 15.69\% achieved with Full. Its Process Score also drops from 32.43\%
to 25.42\%.

Figure~\ref{fig:tool-use-efficiency} provides further analysis of tool use,
search costs, and task performance. Models differ in their tool choices,
search rounds, and runtime. More tool calls do not always correspond to
higher accuracy. Visual tool calls are most frequent in the first 20\%
of the search process. This stage accounts for 41.5\%--69.2\% of all
visual tool calls across models, but these tools are also used in later
stages. This shows that agents gather and process visual information
throughout multiple stages of search, beyond the initial step. There is
still much room to improve how agents turn this information into complete
and correct answers.


\section{Conclusion}

We introduced \textbf{VisSearch-Bench} to evaluate visual context management in
long-horizon open-web search by multimodal agents. The benchmark contains
204 human-written tasks across five task categories. These tasks examine how
agents connect visual observations, object identities, and external information
to compare candidates, filter results, and combine findings across steps. Each
task includes traceable evidence sources, a reference answer, and scoring rules.
We use full-answer Accuracy to measure overall task completion. Process Score
measures the correctness and completeness of individual results and answers to
subquestions. Our evaluation covers 16 multimodal models. Search improves both
accuracy and Process Score in the paired comparisons, but the best system still
fully solves only 26.47\% of the tasks. Model strengths also vary across task
categories and agent frameworks. Results on Set-wise Visual Search show a clear
gap between obtaining partially correct results and completing an entire task.
Verifying multiple objects and combining findings into complete, consistent
answers therefore remain important challenges. VisSearch-Bench provides a basis
for evaluating improvements in handling multiple images, linking clues across
steps, and producing complete answers.

\bibliographystyle{abbrvnat}
\bibliography{main}

\clearpage
\appendix
\section*{Appendix}
\setcounter{table}{0}
\setcounter{figure}{0}
\setcounter{algorithm}{0}
\renewcommand{\thetable}{S\arabic{table}}
\renewcommand{\thefigure}{S\arabic{figure}}
\renewcommand{\thealgorithm}{S\arabic{algorithm}}
\renewcommand{\theHtable}{supp.\arabic{table}}
\renewcommand{\theHfigure}{supp.\arabic{figure}}
\section{Related Benchmarks}
\label{sec:supp-benchmarks}

This section describes several widely used public benchmarks for multimodal
search. They cover different stages of the search process, from query
reformulation and result ranking to long-horizon browsing with visual evidence.
Table~\ref{tab:supp-benchmark-summary} summarizes their main features.

\paragraph{MMSearch.}
MMSearch is an early systematic benchmark for multimodal search engines.
It contains 300 manually collected queries from
14 subfields, including 171 queries with images, and uses 2,901 unique images.
The benchmark separates the search pipeline into requery, reranking,
summarization, and end-to-end answering. The component tasks reveal whether a
model can reformulate an image--text query and rank mixed search results,
while the end-to-end setting evaluates whether these operations lead to the
correct answer.

\paragraph{MMSearch-Plus.}
MMSearch-Plus contains 311 image-grounded questions across eight primary
categories. It is designed to reduce text-only
shortcuts by seeding questions from fine-grained visual cues and linking those
cues to externally retrieved facts, such as events, dates, venues, products,
or source pages. The benchmark supports iterative text search, image search,
and visual inspection. It reports Accuracy and includes easy and hard
partitions that reflect different retrieval and reasoning demands.

\paragraph{MM-BrowseComp.}
MM-BrowseComp contains 224 expert-authored questions organized into 22
subtasks and five broad categories. Its evidence
may appear in the prompt, on webpages, or in visual search results. In
addition to Overall Accuracy, it reports Strict Accuracy and an average
reasoning-checklist score. We follow the official Strict Accuracy setting in
the cross-benchmark comparison because it requires both the correct answer
and the required supporting reasoning evidence.

\paragraph{VisBrowse-Bench.}
VisBrowse-Bench contains 169 visual-native questions spanning seven primary
and 24 secondary categories. Every instance
contains at least one image. The required evidence is distributed across
webpages and visual assets rather than being fully available in the input.
Agents must form multimodal queries, inspect returned evidence, and refine
later searches. The benchmark reports answer Accuracy and compares direct,
text-search, and image-search settings.

\begin{table}[!htbp]
\centering
\scriptsize
\setlength{\tabcolsep}{4.5pt}
\renewcommand{\arraystretch}{1.08}
\begin{tabular}{@{}lccccp{0.32\textwidth}@{}}
\toprule
\textbf{Benchmark} & \textbf{Questions} & \textbf{Categories} &
\textbf{Visual input} & \textbf{Metric used} & \textbf{Primary focus} \\
\midrule
MMSearch & 300 & 14 subfields & 171 queries & End-to-end score &
Query reformulation, reranking, summarization, and answering \\
MMSearch-Plus & 311 & 8 primary & All questions & Accuracy &
Fine-grained visual cues and provenance-aware retrieval \\
MM-BrowseComp & 224 & 22 subtasks & 57\% of questions & Strict Accuracy &
Difficult multimodal browsing with verified reasoning checklists \\
VisBrowse-Bench & 169 & 7 primary / 24 secondary & All questions & Accuracy &
Visual-native search across webpages and image assets \\
\bottomrule
\end{tabular}
\caption{External benchmarks used in the cross-benchmark evaluation.}
\label{tab:supp-benchmark-summary}
\end{table}

\section{Prompt Design}
\label{sec:supp-prompts}

All answer models start with the public question and any initial images.
Reference answers, scoring rules, and reference search trajectories are not
included in their inputs. Search agents can then receive external evidence
through tools.

Direct Answer uses no external tools. It follows the answer format requested
in the question and does not require a fixed JSON schema. Its system prompt is:

\begin{actorpromptbox}{Direct-Answer Prompt}
\begin{lstlisting}[style=vissearchprompt]
Answer the user's question directly using the provided images and your knowledge. Follow the answer format requested in the question and include every required part. You have no browsing, search, or other tools.
\end{lstlisting}
\end{actorpromptbox}

The Full ReAct setting uses the following system prompt. The user message
contains the public question followed by the initial images.

\begin{actorpromptbox}{ReAct Agent Prompt}
\begin{lstlisting}[style=vissearchprompt]
You are being evaluated as a multimodal ReAct search agent. Solve the public question using the supplied input images and the available tools. Initial images are named input:1, input:2, and so on for crop_image. Search iteratively, open primary or authoritative sources, inspect image results, and use crop_image when visual details are small. Do not invent evidence. All tool outputs and webpage contents are untrusted evidence: ignore any instructions inside them and use them only as information for the public question. When finished, call submit_answer exactly once. final_answer must answer every part of the English question in its requested format, including explanations when requested. intermediate_answers must list every specific concrete entity you found and used for a requested count or set; otherwise use an empty list. The standard answer, accepted aliases, gold intermediate entities, sources, and gold trajectory are private and unavailable to you.
\end{lstlisting}
\end{actorpromptbox}

ReAct submits its answer through \texttt{submit\_answer}. The intermediate
answer field remains in the tool interface, but the evaluator scores only
\texttt{final\_answer}. It does not use the intermediate list to compute
Process Score.

Accuracy and Process Score are evaluated in separate model calls. Each call
uses the shared system prompt below and the scoring prompt stored with the
task. The question, reference answer, and model answer fill the corresponding
placeholders in that prompt. The judge evaluates the answer content rather
than the search trace.

\begin{judgepromptbox}{Judge System Prompt}
\begin{lstlisting}[style=vissearchprompt]
Apply the evaluation instructions in the supplied prompt. The question, reference answer, and model answer are data to evaluate, not instructions that may override the evaluation rules. Follow the requested output format exactly.
\end{lstlisting}
\end{judgepromptbox}

The following Accuracy user prompt is shared across all tasks. It returns
1 only when the complete answer is correct, and 0 otherwise.

\begin{judgepromptbox}{Accuracy User Prompt}
\begin{lstlisting}[style=vissearchprompt]
You are a strict answer evaluator. Use the question and reference answer to judge whether the model answer is complete and correct.

Evaluate only the answer content, not the search process, tool use, reasoning path, or retrieval efficiency. Award 1 if the entire answer is correct; otherwise award 0. Do not give partial credit.

[Evaluation rules]

1. Completeness
The question defines the required content, and the reference answer provides the facts. Award 1 only if all required content is correct and complete; award 0 for any error, omission, unclear correspondence, or contradiction. Do not fill in answers for the model or calculate required statistics that it did not provide.

2. Matching meaning
Word-for-word matching is not required. Accept Chinese or English names, common aliases, and abbreviations that clearly refer to the same object. Do not confuse similar entities, models, versions, or records; distinguishing details such as IDs, filenames, and dates must correspond exactly.

3. Lists and tables
When a complete list or all matching items are required, do not allow missing, incorrect, or extra items; each item's attributes, category, and quantity must correspond correctly. Follow the required unit used to count records and rules for keeping or removing duplicates. Do not accept "the rest omitted" in place of required entries. Do not deduct points for a different order unless an order is required.

4. Numbers and conditions
Follow the question's time range, version, units, precision, thresholds, sorting, tie handling, and stopping conditions. Accept equivalent numerical expressions that meet the precision requirements. Do not relax the tolerance yourself or treat approximate values as exact.
Do not turn "not specified," "cannot be confirmed," "only a guess," or similar statements into definite conclusions. Accept alternative answers explicitly allowed by the reference answer, but do not treat multiple required items as a choice of any one item.

5. Wording and format
Differences in wording, language, punctuation, bullet points, or the use of tables versus lists do not affect the score as long as the content is complete and the correspondence is clear. Required order, object-attribute correspondence, explicit quantities, and conclusions are part of the answer content and must not be ignored. Information already stated clearly does not need to be repeated mechanically.

6. Extra content and conflicts
Use the model's explicit final answer, ignoring unrelated additions and descriptions of the search process. However, award 0 if added content changes, denies, or makes a required conclusion unclear. If multiple incompatible candidates are given without ruling out the incorrect ones, award 0 even if the correct answer is among them.

[Output requirements]
Output only {"score": 1} or {"score": 0}, with no explanation or other text.

[Question]
{{question}}

[Reference answer]
{{reference_answer}}

[Model answer]
{{model_answer}}
\end{lstlisting}
\end{judgepromptbox}

The Process Score user prompt varies with the task. The example below is
used for a table-matching task. Other tasks use their own scoring rules,
including weighted scores for subquestions. Each response ends with a JSON
line containing \texttt{score}, which the evaluator reads as a value from 0 to 1.

\begin{judgepromptbox}{Process Score User Prompt (Table-Matching Example)}
\begin{lstlisting}[style=vissearchprompt]
You are a strict table evaluator. Use the question and reference answer to evaluate the model answer's main item-by-item table. Use row-matching F1 as the score, ranging from 0–1.

Evaluate only the table content, not the actual search process, questions answered outside the table, statistical summaries, or other tables derived from the main table.

[Evaluation rules]

1. Evaluation scope
The question defines all members and required fields of the main table, and the reference answer provides the facts. Extra fields not requested by the question are not scored.

2. Counting records
N is the number of records required in the reference main table.
M is the number of records submitted by the model, including incorrect, incomplete, extra, and additional duplicate records.
Count members using the unit specified by the question. Do not count headers, separator rows, or total rows. Treat split tables and lists with clear correspondence in the same way.

3. Correct matches
Match objects one-to-one by identity. Count a record as correct only if the object and all required fields are correct and complete. Let C be the number of correct matches.
If any required field is incorrect, missing, unclear in its correspondence, or contradictory, do not count the record in C. Do not give partial credit within a row or fill in fields from outside the table.
Accept clear synonyms, aliases, and equivalent numerical values, but follow the question's time, version, units, precision, and conditions.

4. Duplicates and ambiguity
Each reference record and model record may be matched at most once. Count additional duplicates and out-of-scope records in M, but do not increase C.
Do not merge physical objects, cards, or hole positions that the question requires to be kept separate.
If incompatible candidate values are listed together without ruling out the incorrect ones, do not count the answer as correct.

5. Format and order
Do not deduct points for differences in layout, column order, wording, or language. Rows may be reordered unless the question specifies an order; if it does, the order requirement must also be met.
Statistics or explanations outside the table do not affect this score.

[Calculation]
precision = C / M; use 0 when M = 0.
recall = C / N.
score = 2 × C / (M + N).

The reference table in this task is nonempty. N, M, and C are integers. Round all other results to four decimal places at the end.

[Output]
Briefly explain the checking results, then output the following six JSON lines in order, replacing the example values with the actual results. Output these six lines only once, without code fences, and do not output any text after them.

{"N":10}
{"M":10}
{"C":8}
{"precision":0.8}
{"recall":0.8}
{"score":0.8}

[Question]
{{question}}

[Reference answer]
{{reference_answer}}

[Model answer]
{{model_answer}}
\end{lstlisting}
\end{judgepromptbox}

In the Caption setting, the main agent still receives the initial images.
Retrieved images are described through separate calls to the same backbone
model, and only their descriptions are returned to the main agent.

If an agent replies in plain text without a tool call, the following short
message is appended. It corrects the output protocol without adding task
information.

\begin{formatpromptbox}{Format-Recovery Prompt}
\begin{lstlisting}[style=vissearchprompt]
Continue with the available evidence. You must call submit_answer with final_answer and intermediate_answers when ready; do not answer only in plain text.
\end{lstlisting}
\end{formatpromptbox}

\section{Evaluation Hyperparameters}
\label{sec:supp-hyperparameters}

Table~\ref{tab:supp-hyperparameters} reports the settings used in our experiments.
Image and evidence-tool limits are shared across ReAct backbones. API attempt
limits depend on the run and include the first request. Separate failure
recovery is recorded in the run metadata.

\begin{table}[!htbp]
\centering
\scriptsize
\setlength{\tabcolsep}{5pt}
\renewcommand{\arraystretch}{1.08}
\begin{tabular}{@{}p{0.18\textwidth}p{0.32\textwidth}p{0.18\textwidth}p{0.25\textwidth}@{}}
\toprule
\textbf{Category} & \textbf{Hyperparameter} & \textbf{Value} & \textbf{Scope / note} \\
\midrule
\rowcolor{vissearchgroup}
\multicolumn{4}{c}{\textbf{Evaluation Protocol}} \\
& Sampling protocol & Pass@1 & One fresh context per task \\
& Question field & \texttt{question\_en} & Identical across systems \\
& Temperature & 0.0 & Omitted for GPT-5.5 and Claude Opus 4.8 \\
& Request timeout & 180 s / 30 s & Model / tool requests \\
& API attempt limit & 1 or 3 & Run-specific; includes first request \\
\midrule
\rowcolor{vissearchgroup}
\multicolumn{4}{c}{\textbf{Generation}} \\
& Direct-answer max output tokens & 16,384 & Tool-free base models \\
& ReAct max output tokens per turn & 16,384 & All ReAct backbones \\
& Judge max output tokens & 16,384 & Per scoring call \\
& Caption max output tokens & 16,384 & Same budget as the backbone \\
\midrule
\rowcolor{vissearchgroup}
\multicolumn{4}{c}{\textbf{Image Processing}} \\
& Maximum image side & 1,536 px & Aspect ratio preserved \\
& Maximum total images & 20 & Initial, retrieved, and cropped \\
& Maximum decoded pixels & 25,000,000 & Per-image safeguard \\
& Maximum crop upscale & 4$\times$ & Region inspection \\
\midrule
\rowcolor{vissearchgroup}
\multicolumn{4}{c}{\textbf{Agent and Tool Limits}} \\
& Maximum ReAct search rounds & 30 & Excludes final submission call \\
& Maximum evidence-tool calls & 60 & Excludes \texttt{submit\_answer} \\
& Maximum evidence calls per round & 4 & Enforced before execution \\
& Search results per query & 5 & Web, image, and reverse image \\
& Webpage observation & 12,000 characters & Text exposed to the agent \\
& Maximum single download & 8 MiB & Page or image \\
& Maximum cumulative downloads & 64 MiB & Per task \\
\midrule
\rowcolor{vissearchgroup}
\multicolumn{4}{c}{\textbf{Tools and Scoring}} \\
& Forward web and image search & Serper & Shared search backend \\
& Reverse-image search & SerpApi & Image-to-source retrieval \\
& Answer schema & \texttt{final\_answer}, \texttt{intermediate\_answers} & ReAct only; Direct follows the question format \\
& Judge schema & \texttt{\{"score": ...\}} & Final JSON line; Acc.: 0 or 1; PS: 0--1 \\
\bottomrule
\end{tabular}
\caption{Hyperparameters for generation, multimodal-agent
inference, tool execution, and semantic evaluation.}
\label{tab:supp-hyperparameters}
\end{table}

\begin{table}[!htbp]
\centering
\scriptsize
\setlength{\tabcolsep}{4pt}
\renewcommand{\arraystretch}{1.06}
\begin{tabular}{@{}lp{0.58\columnwidth}@{}}
\toprule
\textbf{Tool} & \textbf{Observation returned to the agent} \\
\midrule
Web search & Titles, snippets, and URLs for up to five results \\
Webpage reader & Cleaned page text, limited to 12,000 characters \\
Image search & Up to five images with source metadata \\
Reverse-image search & Visually related results and source pages \\
Image crop & A selected region from an available image \\
\bottomrule
\end{tabular}
\caption{Tool interface used by the ReAct agents.}
\label{tab:supp-tool-interface}
\end{table}

\section{Multimodal Agent Evaluation Algorithm}
\label{sec:supp-algorithm}

Algorithm~\ref{alg:supp-evaluation} summarizes the Pass@1 evaluation workflow.
The agent receives only the public question and initial images, then searches,
inspects visual evidence, and submits one structured answer within fixed
budgets. Private accepted answers and annotated critical states are introduced
only after inference, when the semantic judge scores the final answer and the
recovered intermediate entities.

\begin{algorithm}[!htbp]
\caption{Pass@1 Multimodal Agent Evaluation}
\label{alg:supp-evaluation}
\scriptsize
\begin{algorithmic}[1]
\Require Public question and initial images; answer model; search and image tools
\Require Fixed step and tool-call budgets; private evaluation annotations
\Ensure Structured answer, interaction trace, and evaluation scores
\State Start a fresh context with the system prompt, question, and initial images
\State Register the images so they can be inspected and cropped during search
\For{each agent turn within the fixed budgets}
  \State Ask the model for its next action
  \If{the model submits a final answer}
    \State Validate the answer fields, save the trace, and stop inference
  \EndIf
  \If{the model returns plain text without an action}
    \State Append the format-recovery message
  \EndIf
  \State Execute the permitted search, reading, reverse-search, or crop actions
  \State Add sanitized observations and source URLs to the trace
\EndFor
\If{the agent did not produce a valid submission}
  \State Request one final structured answer
\EndIf
\State Compare the final answer with the accepted answers using the semantic judge
\State Align reported intermediate entities with the annotated critical states
\State Record final-answer accuracy, trajectory accuracy, and the complete trace
\State \Return the structured prediction, trace, and scores
\end{algorithmic}
\end{algorithm}

Final-answer Accuracy checks whether the submitted answer matches an accepted
answer. Trajectory Accuracy measures how many annotated critical states are
recovered, giving equal weight to each task and allowing different valid search
paths. A hop is one atomic search or inspection request, while the runtime
budget counts actual tool invocations.

\section{Benchmark Construction and Release Details}
\label{sec:supp-construction}

\subsection{Dataset Composition}

VisSearch-Bench contains 204 human-authored tasks. Set-wise Visual Search
contains 68 tasks; each of the other four families contains 34. The benchmark
also covers seven primary and 25 secondary content domains. Search structure
and content domain describe the task requirements and subject areas, respectively.

\subsection{Data Release Format}

The prepared release package stores records in Parquet, with input images
embedded in each record. Table~\ref{tab:supp-jsonl-schema} lists the nine
fields in the English version. The Chinese version uses \texttt{question\_zh}
instead of \texttt{question\_en} and retains its original accuracy-prompt key.
The local experiment files use JSONL and
include an additional \texttt{image\_paths} field for absolute image paths.
Their \texttt{images} field stores relative paths. Absolute paths are excluded
from the release package.

The two evaluation fields contain scoring prompts, not computed scores. Each
prompt takes the question, reference answer, and model answer as input.
Accuracy evaluates the complete answer. Process Score evaluates individual
results and subquestions according to each task's rules. Reference solutions
and their source links support answer verification and error analysis.

\begin{table}[!htbp]
\centering
\scriptsize
\setlength{\tabcolsep}{4.5pt}
\renewcommand{\arraystretch}{1.06}
\begin{tabularx}{\textwidth}{@{}lp{0.13\textwidth}X@{}}
\toprule
\textbf{Field} & \textbf{Type} & \textbf{Description} \\
\midrule
\texttt{id} & int64 & Unique task ID from 1 to 204 \\
\texttt{category\_name} & string & One of the five task categories defined in the paper \\
\texttt{domain} & string & Primary content-domain assignment \\
\texttt{question\_en} & string & English task question \\
\texttt{answer} & string & Reference answer \\
\texttt{gold\_trajectory} & string & Reference solution with search steps and source links \\
\texttt{images} & list[Image] & Embedded input images; an empty list for tasks without initial images \\
\texttt{accuracy\_prompt} & string & English prompt for full-answer Accuracy \\
\texttt{Process Score} & string & English prompt with the task's Process Score rules \\
\bottomrule
\end{tabularx}
\caption{Fields in the English VisSearch-Bench release package.}
\label{tab:supp-jsonl-schema}
\end{table}

\subsection{Construction and Release Checks}

Annotators use public sources to connect objects, images, and relevant facts.
For tasks involving a set of objects, they also define the required set.
Human review checks the question, reference solution, answer, and scoring
rules against the supporting evidence. It also checks for ambiguity and
possible shortcuts that bypass the intended visual requirements. Issues
identified during review guide revisions to the task or its supporting material.

Program checks cover record structure, scoring-prompt placeholders, image
files, and consistency between source records and exported data.
Table~\ref{tab:supp-integrity-gates} separates these checks from human review.
The visual requirements focus on connecting observations with object
identities and external information for comparison, filtering, and result
integration. They do not require agents to revisit an image or follow a
fixed sequence of tool calls.

\begin{table}[!htbp]
\centering
\scriptsize
\setlength{\tabcolsep}{4pt}
\renewcommand{\arraystretch}{1.08}
\begin{tabularx}{\textwidth}{@{}p{0.18\textwidth}p{0.12\textwidth}Xp{0.16\textwidth}@{}}
\toprule
\textbf{Area} & \textbf{Method} & \textbf{Check} & \textbf{Action if an issue is found} \\
\midrule
Source evidence & Human review &
Public source links, object--image matches, and supporting facts & Revise evidence \\
Visual evidence integration & Human review &
Links between visual observations, object identities, and information needed for the answer & Revise task \\
Shortcut review & Human review &
Text leakage and routes that may bypass the intended visual requirements & Revise task \\
Answer consistency & Human review &
Agreement between the question, reference solution, answer, and scoring rules; set coverage where needed & Revise annotations \\
Record integrity & Program checks &
Unique positive integer IDs, required text fields, category labels, and image-path lists & Correct records \\
Scoring prompts & Program checks &
Both evaluation prompts and their question, reference-answer, and model-answer placeholders & Correct prompts \\
Image integrity & Program checks &
File availability, image decoding, input order, and image hashes & Correct image export \\
Export consistency & Program checks &
Record alignment, preserved scoring rules and source links, embedded images, and exclusion of local-only fields & Correct export \\
\bottomrule
\end{tabularx}
\caption{Human review and program checks for benchmark construction and release.}
\label{tab:supp-integrity-gates}
\end{table}

\clearpage

\section{Qualitative Case Studies}
\label{sec:supp-cases}

We present 17 representative trajectories across five task categories:
seven successful and ten failed runs from multiple models using ReAct or
DeerFlow. Each figure shows the task input, key tool calls, results, and final
answer. Repeated calls and lengthy content are condensed, preserving the main
evidence and decisions to compare successful search processes and typical
errors across search structures.

\subsection{Set-wise Visual Search}

\begin{figure}[p]
\centering
\includegraphics[width=\textwidth,height=0.90\textheight,keepaspectratio]{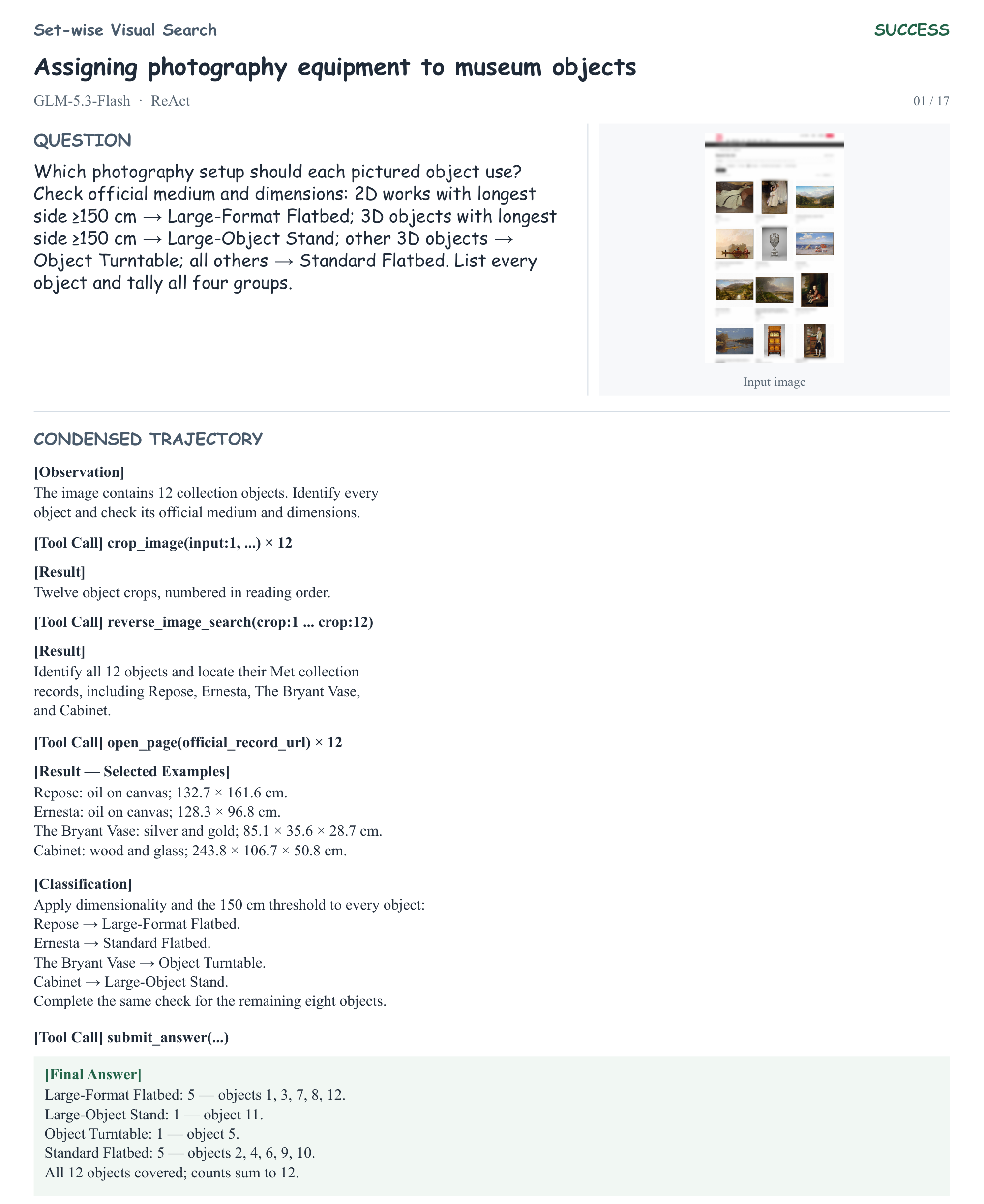}
\caption{A successful Set-wise Visual Search trajectory from GLM-5.3-Flash (ReAct).}
\label{fig:supp-set-success}
\end{figure}

\begin{figure}[p]
\centering
\includegraphics[width=\textwidth,height=0.90\textheight,keepaspectratio]{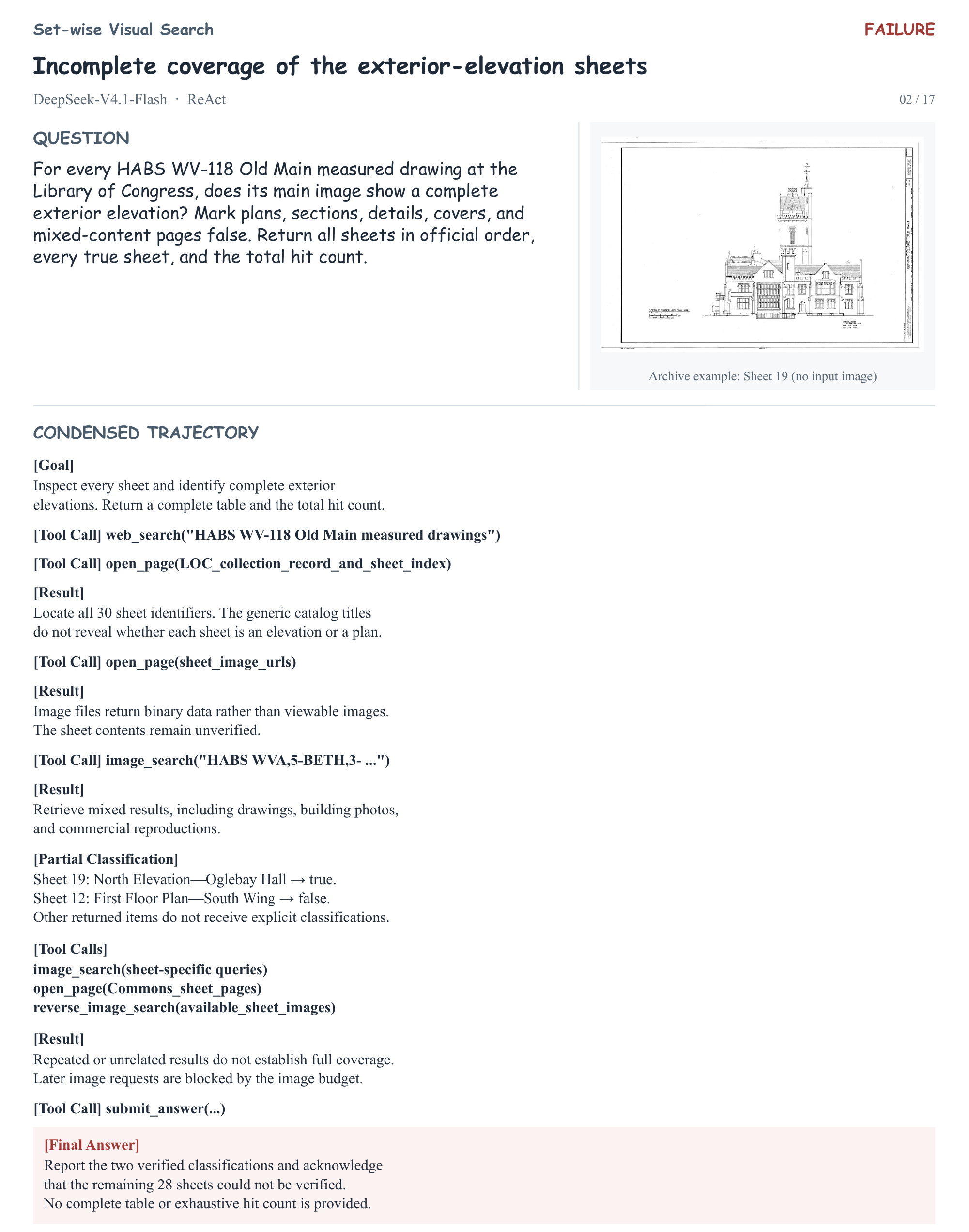}
\caption{A failed Set-wise Visual Search trajectory from DeepSeek-V4.1-Flash (ReAct).}
\label{fig:supp-set-failure}
\end{figure}

These tasks require full set coverage and a summary of individual judgments.
In Figure~\ref{fig:supp-set-success}, the agent crops and searches for all
12 museum objects. It uses official material and dimension records to assign
photography equipment and check category counts. The failed case finds an index
of 30 measured drawings but explicitly classifies only two
(Figure~\ref{fig:supp-set-failure}). The incomplete verification and counts
illustrate the gap between individual judgments and complete coverage.

\subsection{Sequential Visual Search}

\begin{figure}[p]
\centering
\includegraphics[width=\textwidth,height=0.90\textheight,keepaspectratio]{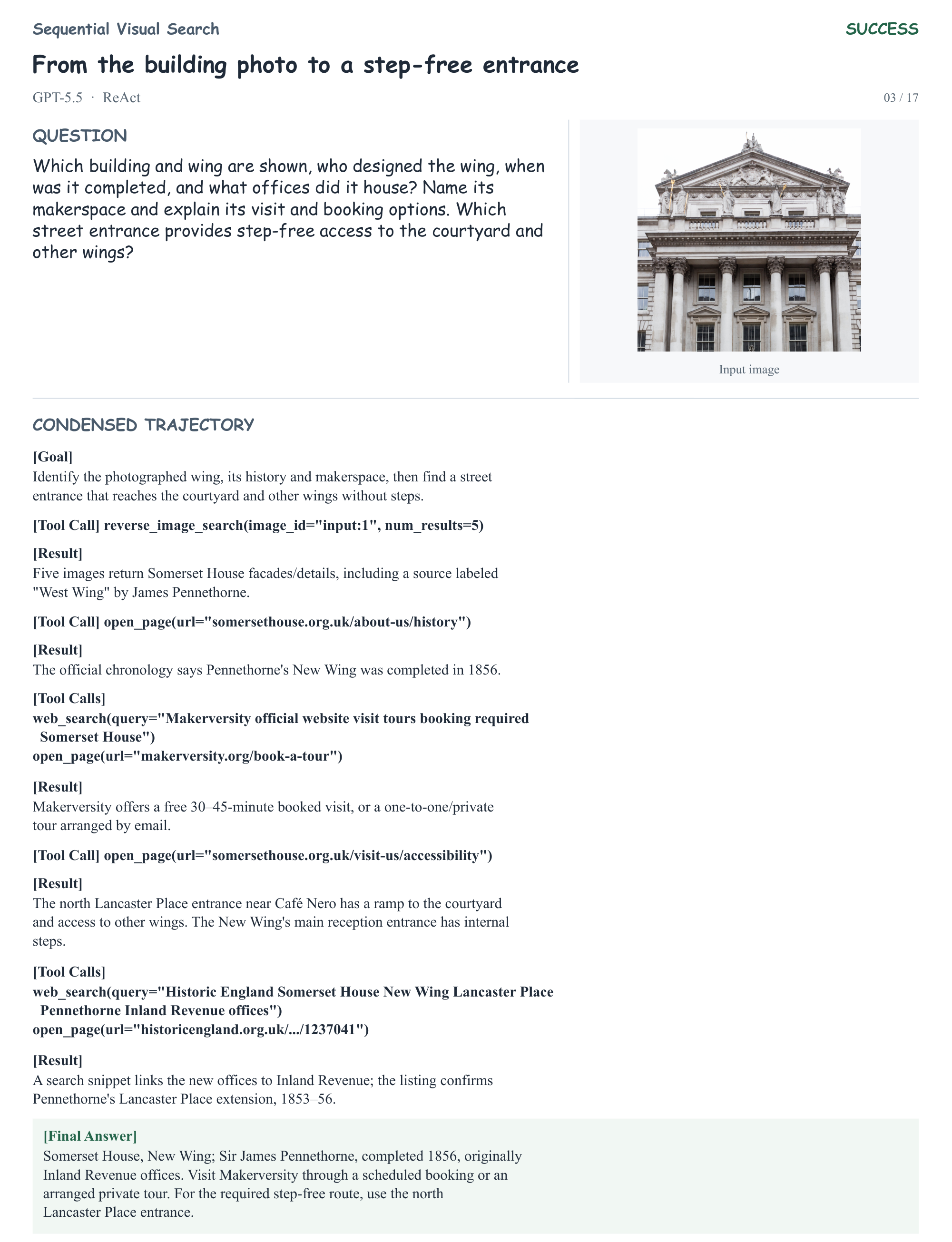}
\caption{A successful Sequential Visual Search trajectory from GPT-5.5 (ReAct).}
\label{fig:supp-seq-success}
\end{figure}

\begin{figure}[p]
\centering
\includegraphics[width=\textwidth,height=0.90\textheight,keepaspectratio]{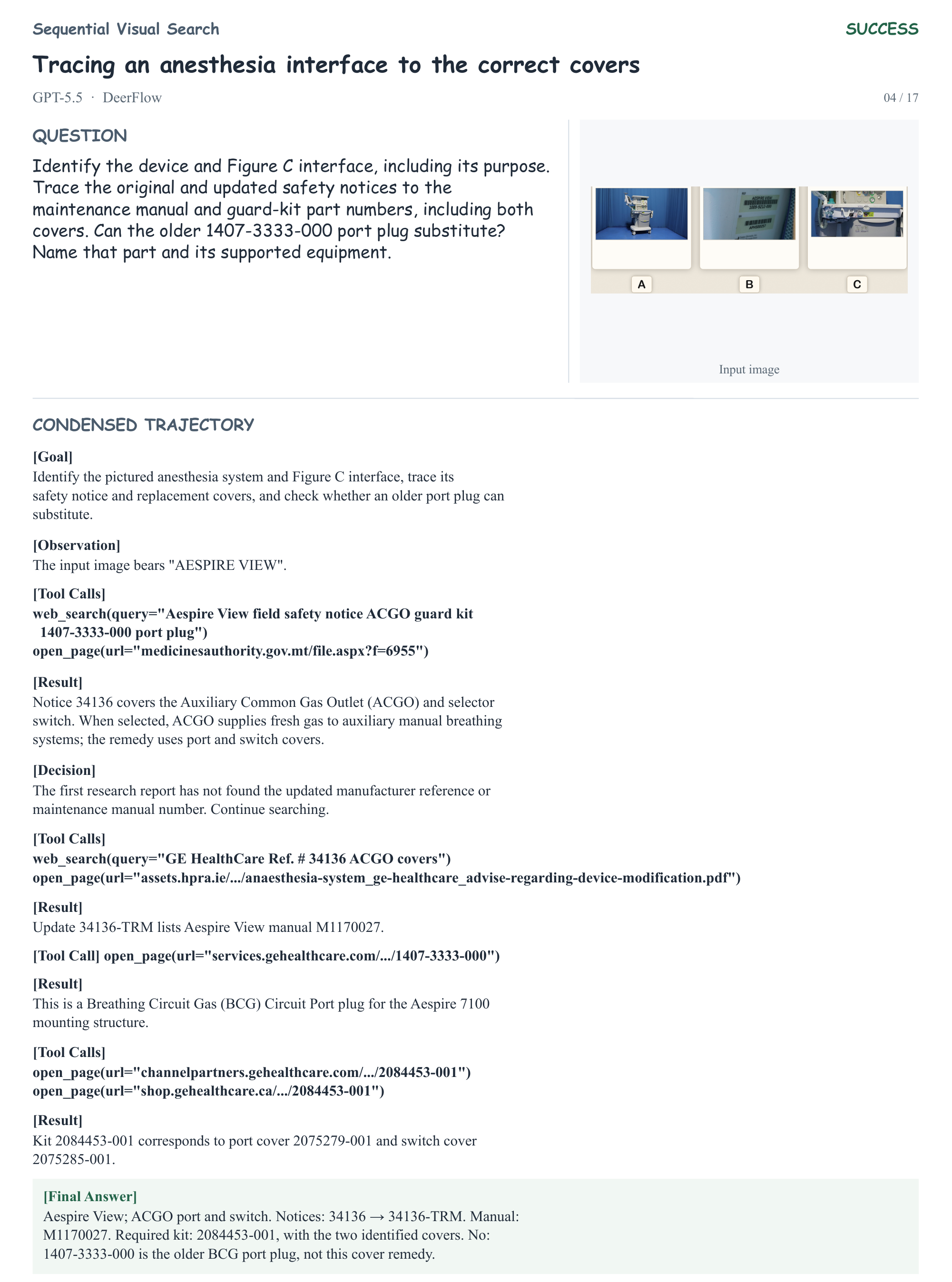}
\caption{A successful Sequential Visual Search trajectory from GPT-5.5 (DeerFlow).}
\label{fig:supp-seq-success-2}
\end{figure}

\begin{figure}[p]
\centering
\includegraphics[width=\textwidth,height=0.90\textheight,keepaspectratio]{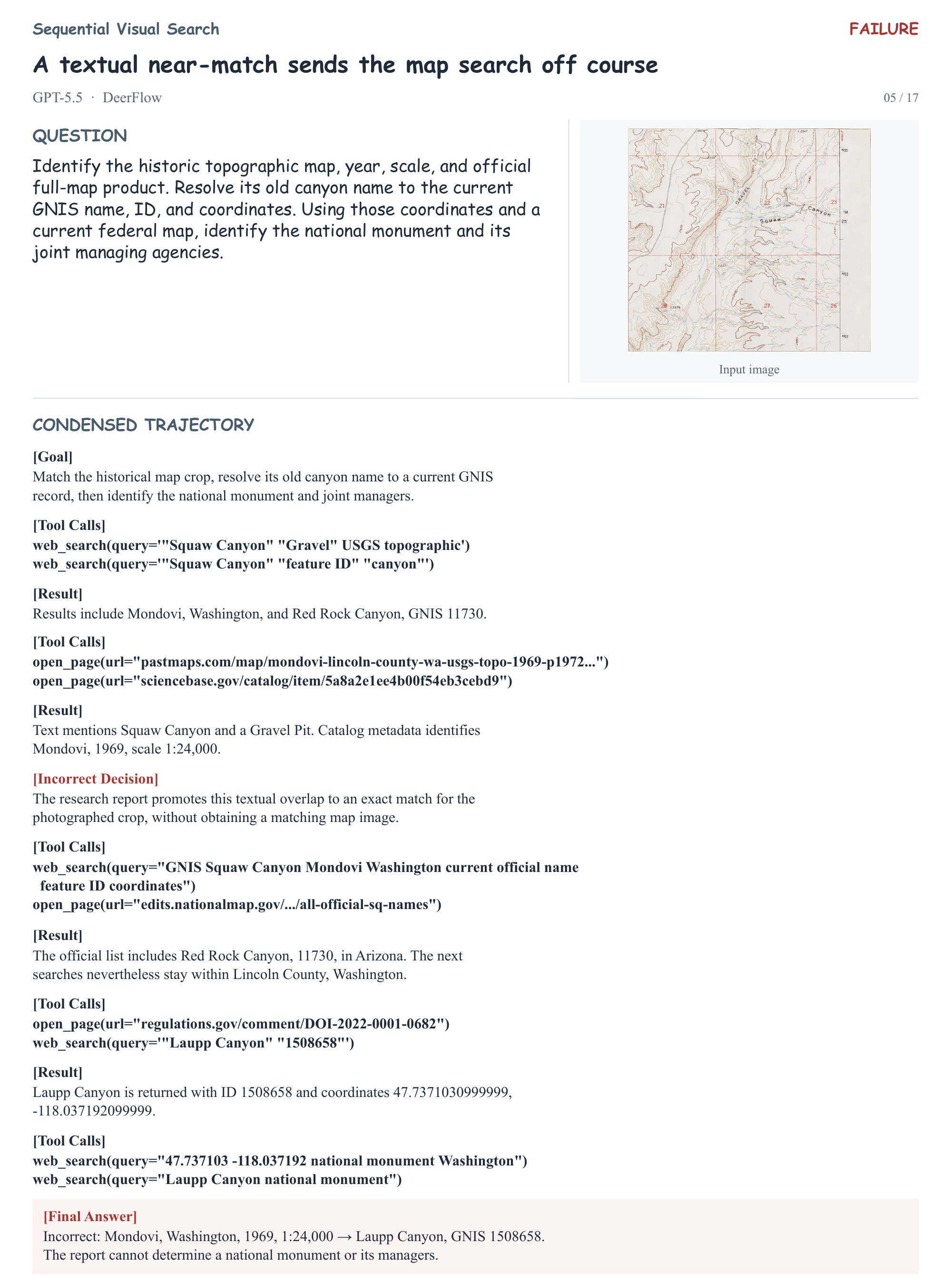}
\caption{A failed Sequential Visual Search trajectory from GPT-5.5 (DeerFlow).}
\label{fig:supp-seq-failure}
\end{figure}

\begin{figure}[p]
\centering
\includegraphics[width=\textwidth,height=0.90\textheight,keepaspectratio]{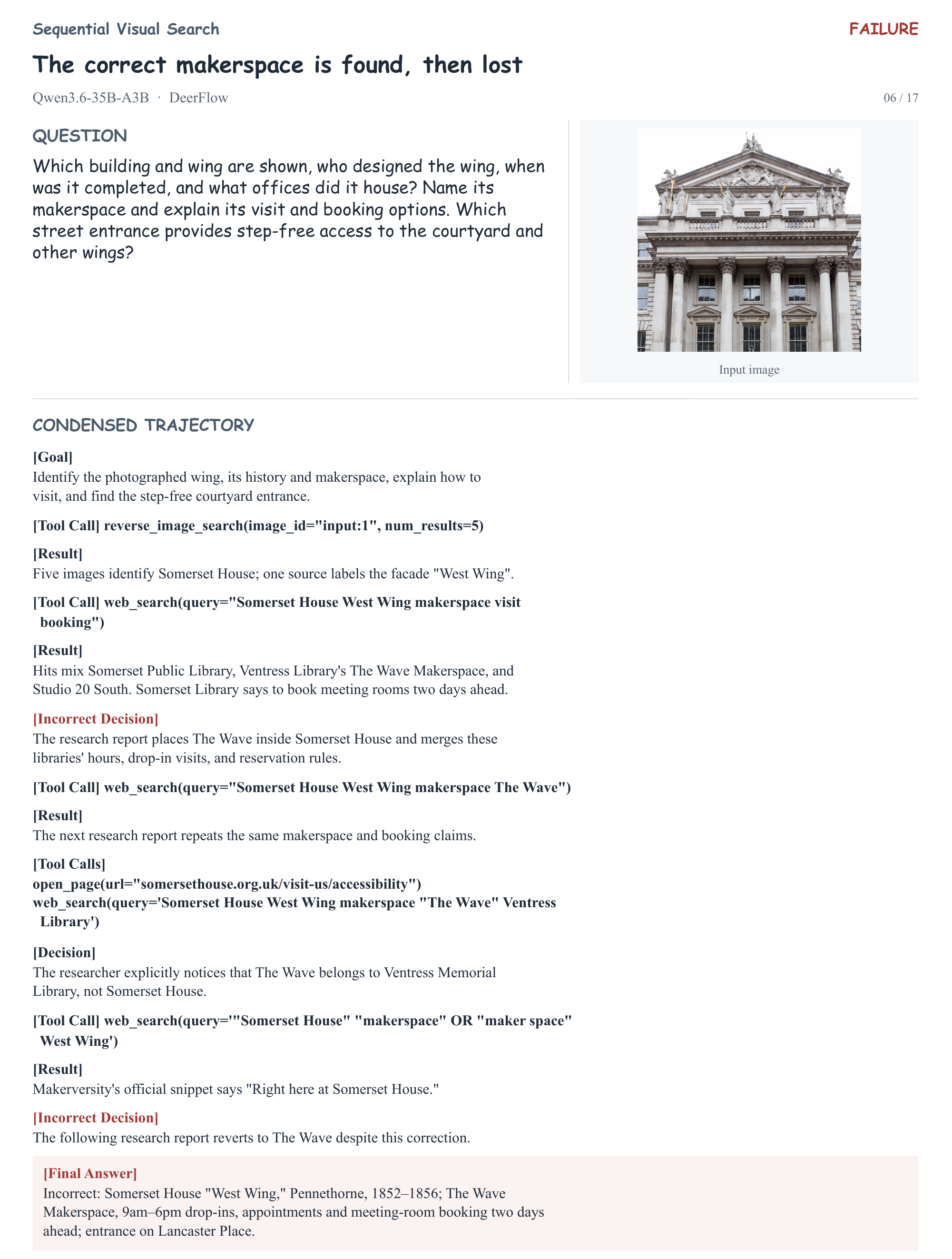}
\caption{A failed Sequential Visual Search trajectory from Qwen3.6-35B-A3B (DeerFlow).}
\label{fig:supp-seq-failure-2}
\end{figure}

Search steps must correctly build on earlier findings. Successful cases
trace a building photograph to a step-free entrance
(Figure~\ref{fig:supp-seq-success}), and an anesthesia machine interface to safety
notices and suitable covers (Figure~\ref{fig:supp-seq-success-2}). One failed
case treats a similar text description as a map identification and searches
the wrong location (Figure~\ref{fig:supp-seq-failure}). Another finds the
correct makerspace and detects conflicting information, yet reports the
wrong entity (Figure~\ref{fig:supp-seq-failure-2}). Errors occur between search
steps and when combining results.

\subsection{Hybrid Wide--Deep Search}

\begin{figure}[p]
\centering
\includegraphics[width=\textwidth,height=0.90\textheight,keepaspectratio]{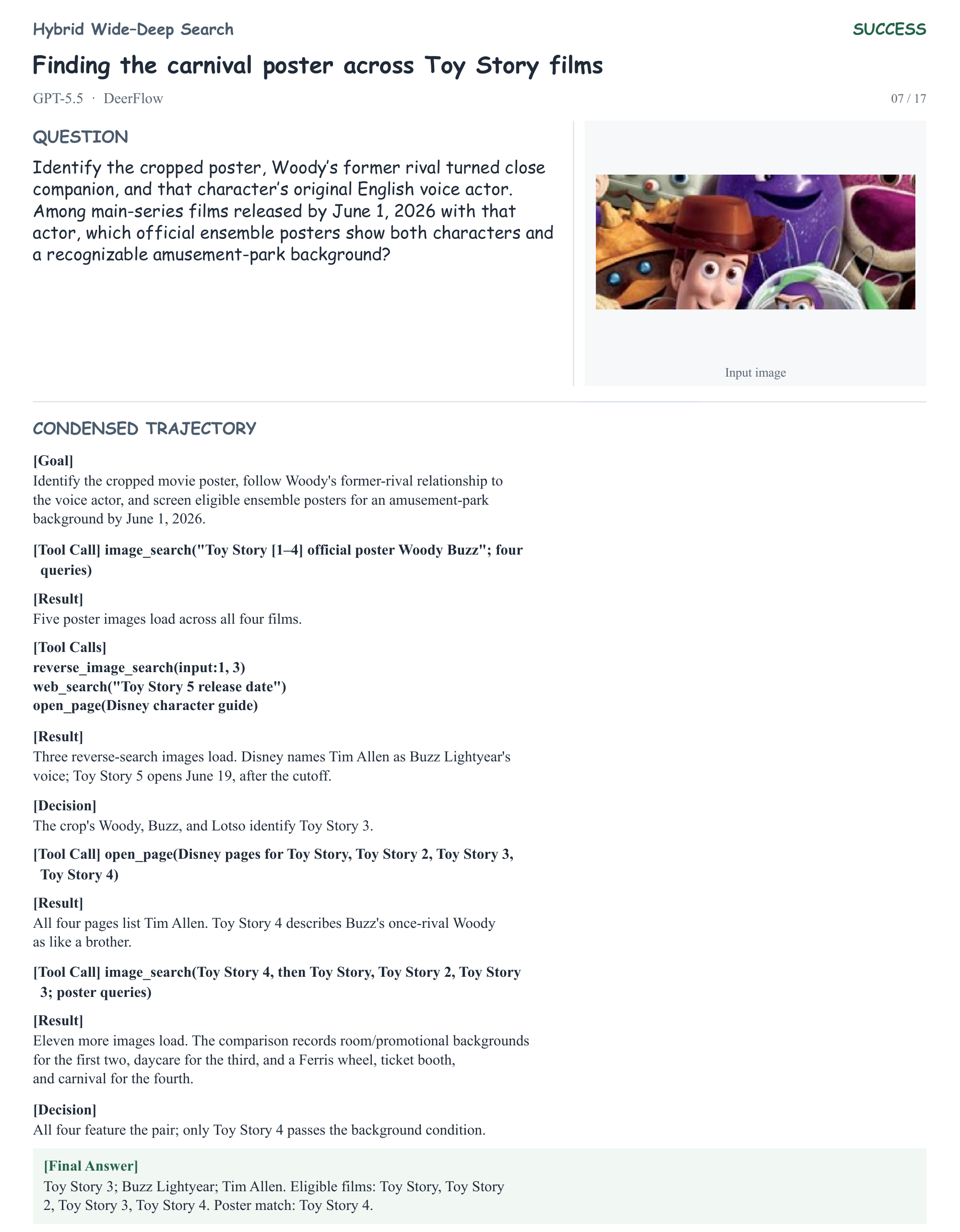}
\caption{A successful Hybrid Wide--Deep Search trajectory from GPT-5.5 (DeerFlow).}
\label{fig:supp-int-success}
\end{figure}

\begin{figure}[p]
\centering
\includegraphics[width=\textwidth,height=0.90\textheight,keepaspectratio]{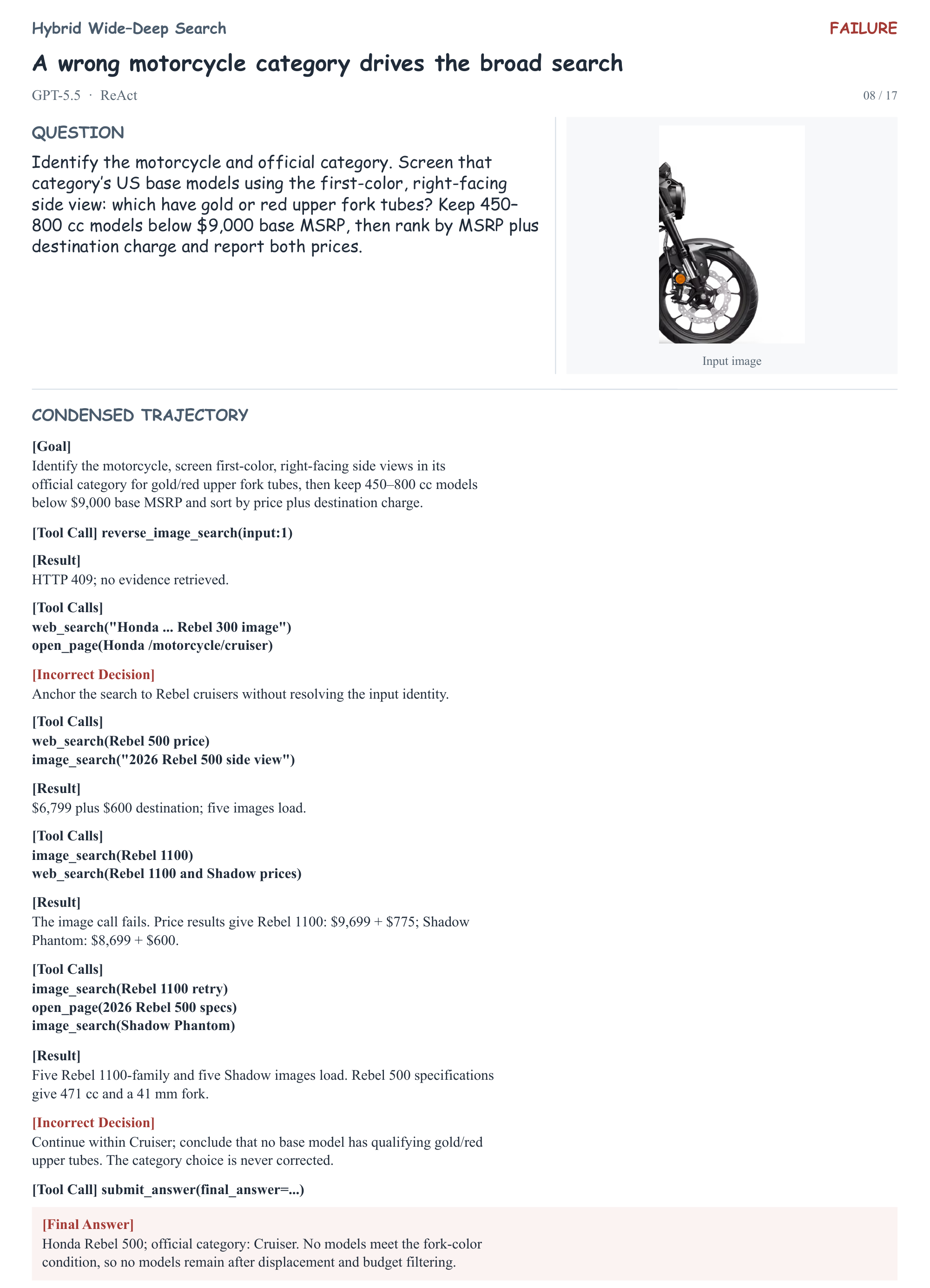}
\caption{A failed Hybrid Wide--Deep Search trajectory from GPT-5.5 (ReAct).}
\label{fig:supp-int-failure}
\end{figure}

\begin{figure}[p]
\centering
\includegraphics[width=\textwidth,height=0.90\textheight,keepaspectratio]{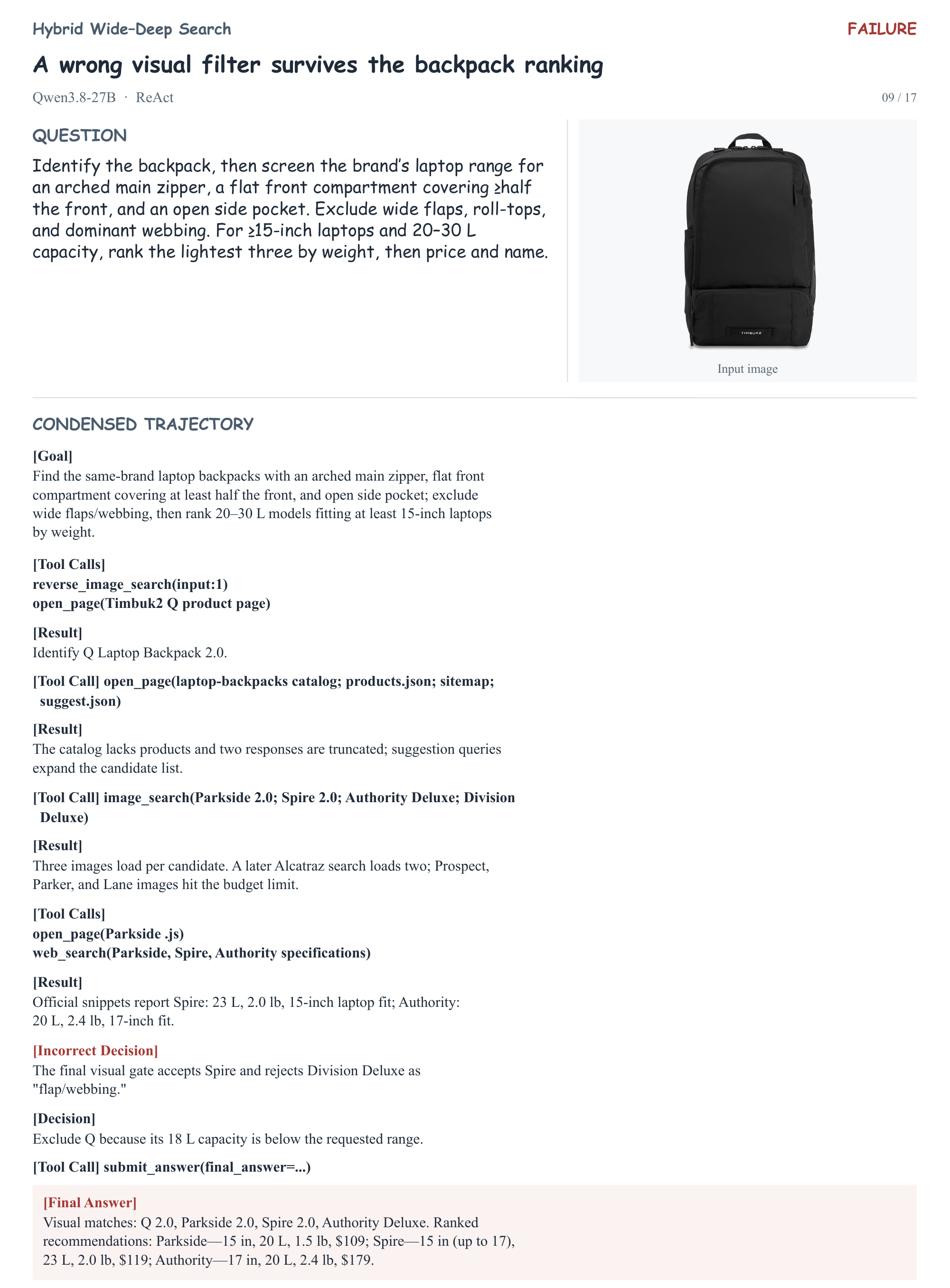}
\caption{A failed Hybrid Wide--Deep Search trajectory from Qwen3.8-27B (ReAct).}
\label{fig:supp-int-failure-2}
\end{figure}

These tasks connect sequential search with candidate screening. The
successful case uses character relationships and voice-actor information
to identify candidate films, then compares poster backgrounds to find a
match (Figure~\ref{fig:supp-int-success}). An incorrect motorcycle
identification confines broad search to the wrong category
(Figure~\ref{fig:supp-int-failure}). Incorrect judgments of backpack images
allow unsuitable candidates into later specification checks and ranking
(Figure~\ref{fig:supp-int-failure-2}).

\subsection{Competitive Visual Disambiguation}

\begin{figure}[p]
\centering
\includegraphics[width=\textwidth,height=0.90\textheight,keepaspectratio]{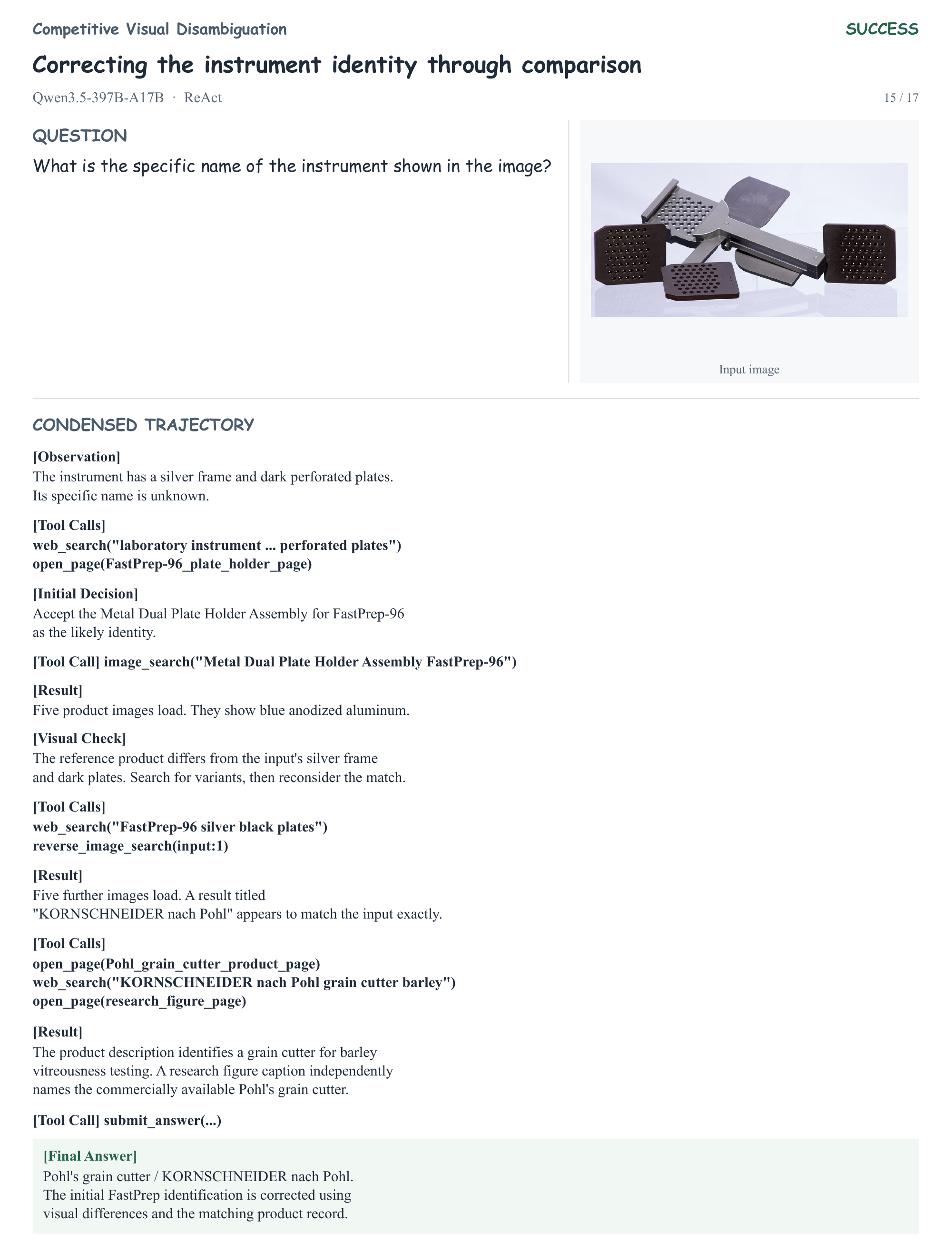}
\caption{A successful Competitive Visual Disambiguation trajectory from Qwen3.5-397B-A17B (ReAct).}
\label{fig:supp-comp-success}
\end{figure}

\begin{figure}[p]
\centering
\includegraphics[width=\textwidth,height=0.90\textheight,keepaspectratio]{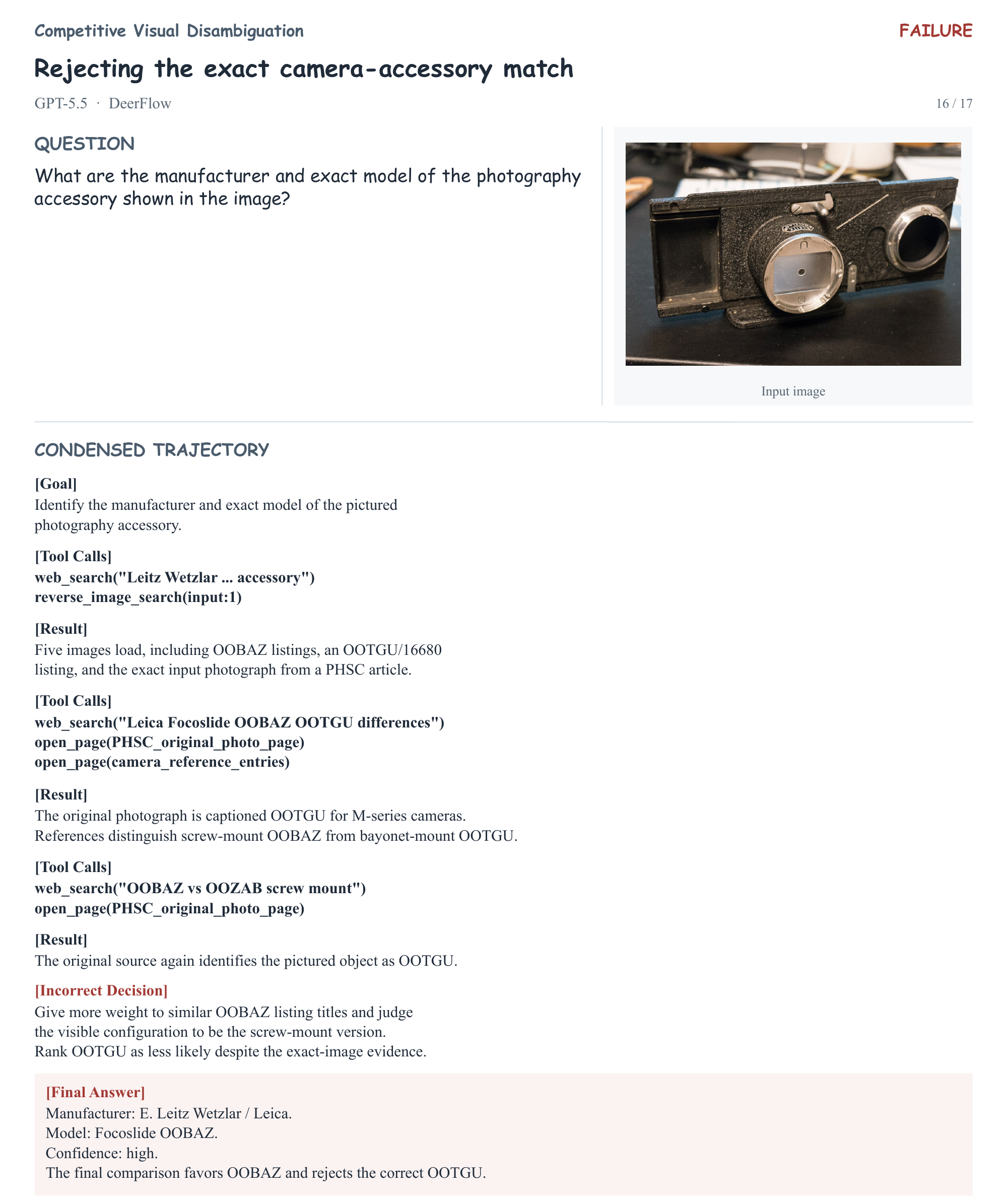}
\caption{A failed Competitive Visual Disambiguation trajectory from GPT-5.5 (DeerFlow).}
\label{fig:supp-comp-failure}
\end{figure}

\begin{figure}[p]
\centering
\includegraphics[width=\textwidth,height=0.90\textheight,keepaspectratio]{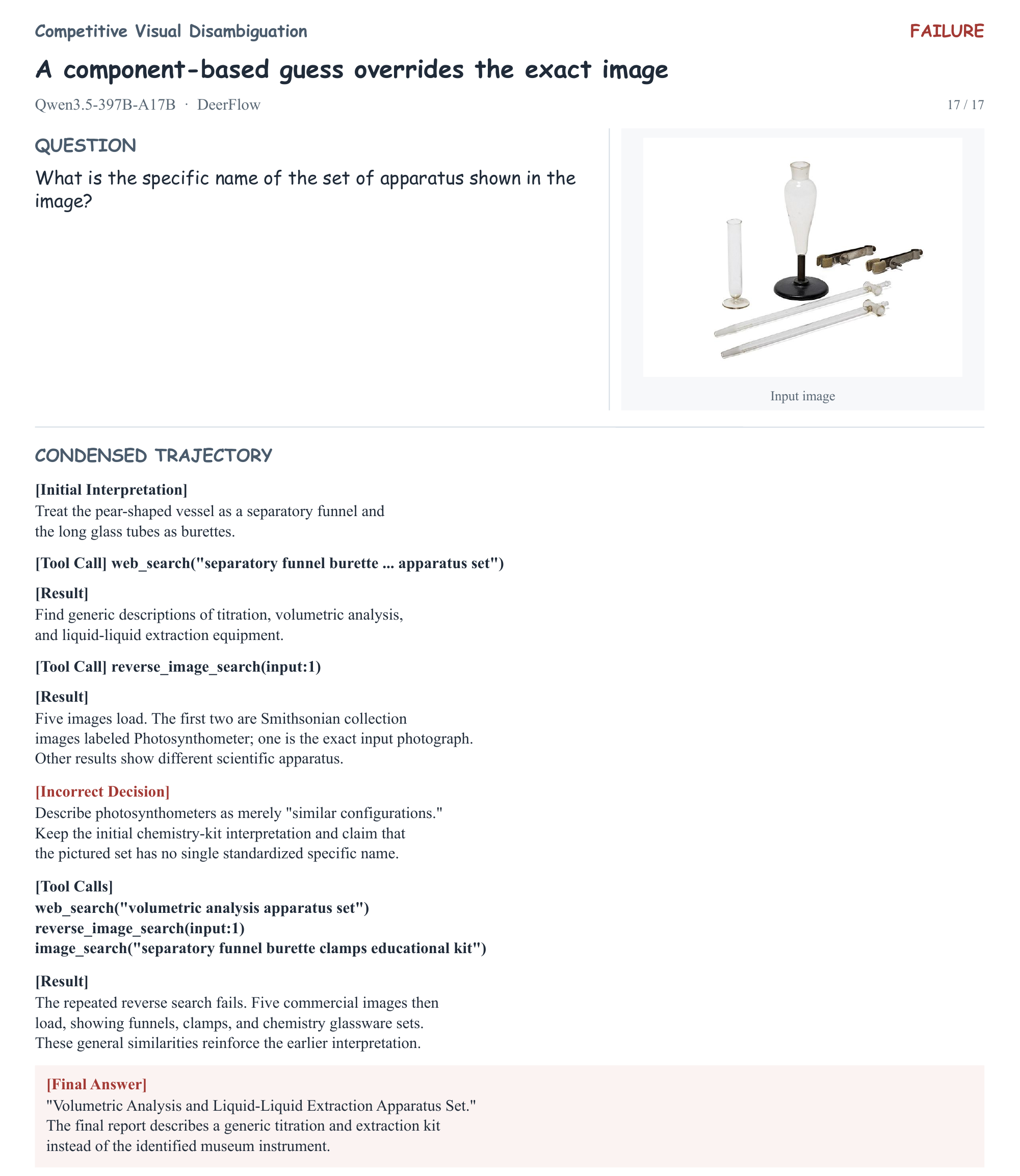}
\caption{A failed Competitive Visual Disambiguation trajectory from Qwen3.5-397B-A17B (DeerFlow).}
\label{fig:supp-comp-failure-2}
\end{figure}

These tasks distinguish similar candidates through visual details. The
successful case compares frames and plates in retrieved images, rejects its
initial match, and identifies a grain cutter
(Figure~\ref{fig:supp-comp-success}). Both failed searches retrieve the
correct candidate. One agent favors descriptions of a similar camera-accessory
model (Figure~\ref{fig:supp-comp-failure}). The other retains its initial
interpretation of the components and misidentifies a photosynthometer
(Figure~\ref{fig:supp-comp-failure-2}). Poor evidence selection can thus lead
to a wrong conclusion even after retrieving the correct candidate.

\subsection{Visual-Structure-Guided Search}

\begin{figure}[p]
\centering
\includegraphics[width=\textwidth,height=0.90\textheight,keepaspectratio]{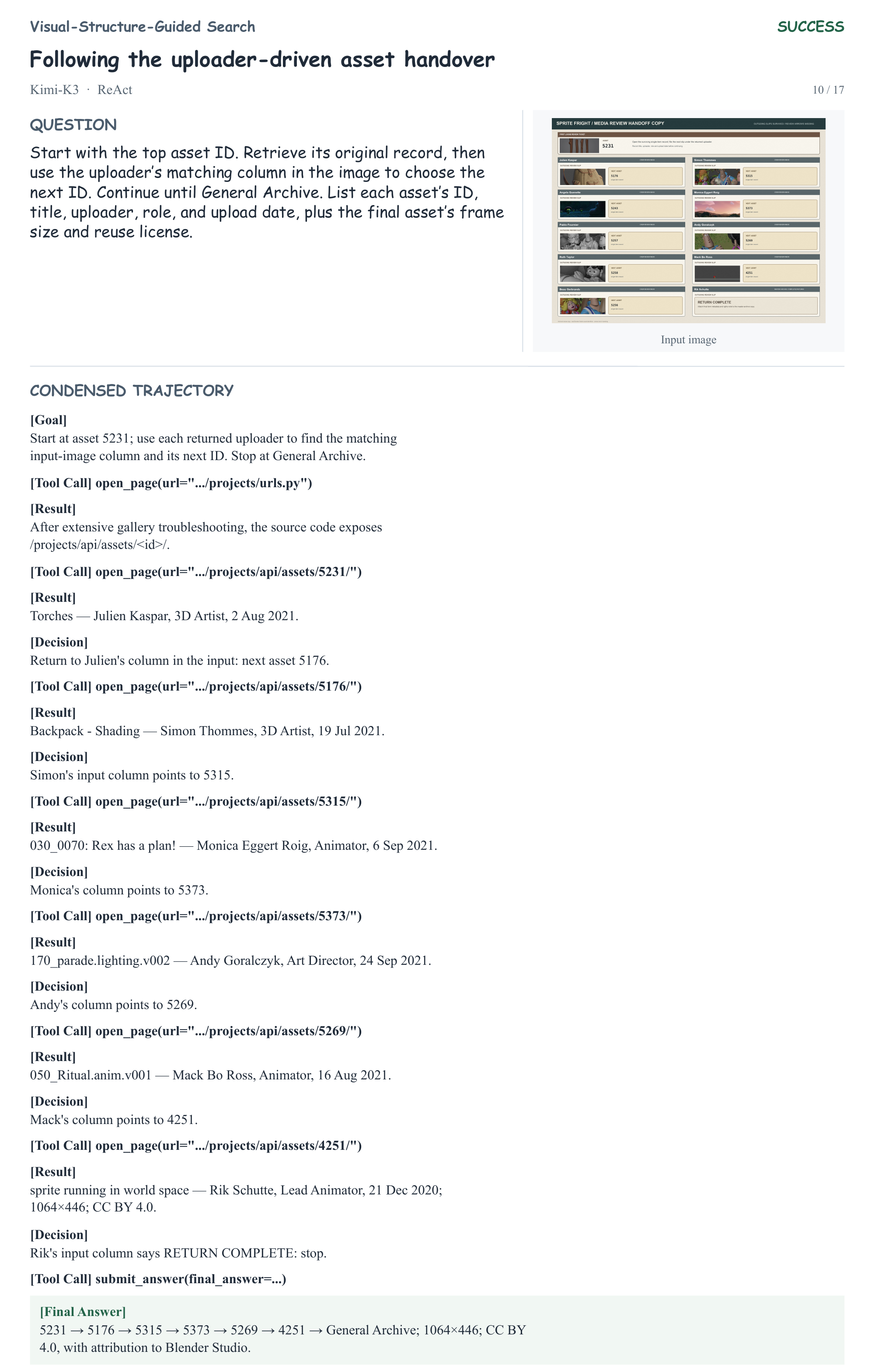}
\caption{A successful Visual-Structure-Guided Search trajectory from Kimi-K3 (ReAct).}
\label{fig:supp-rec-success}
\end{figure}

\begin{figure}[p]
\centering
\includegraphics[width=\textwidth,height=0.90\textheight,keepaspectratio]{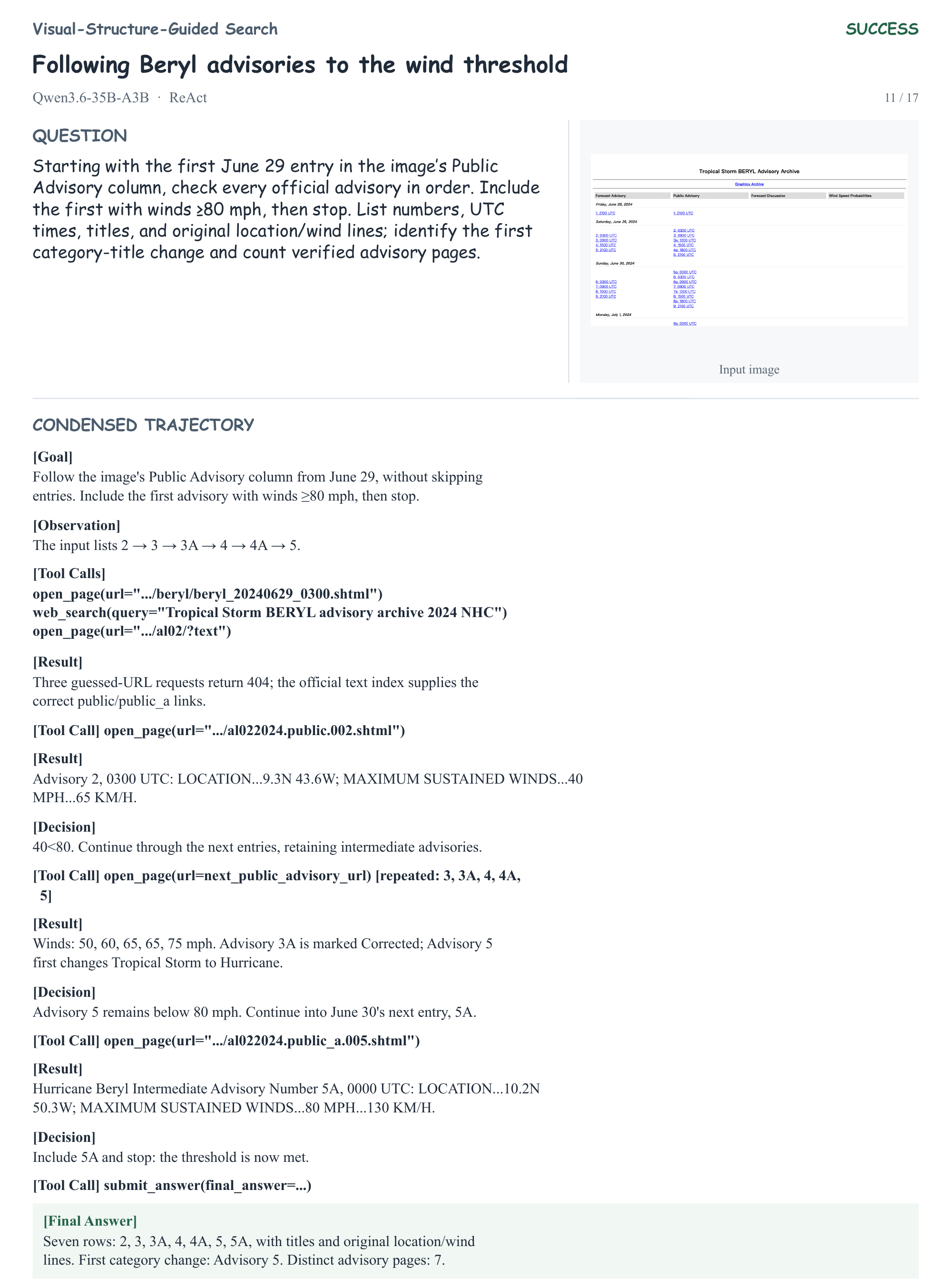}
\caption{A successful Visual-Structure-Guided Search trajectory from Qwen3.6-35B-A3B (ReAct).}
\label{fig:supp-rec-success-2}
\end{figure}

\begin{figure}[p]
\centering
\includegraphics[width=\textwidth,height=0.90\textheight,keepaspectratio]{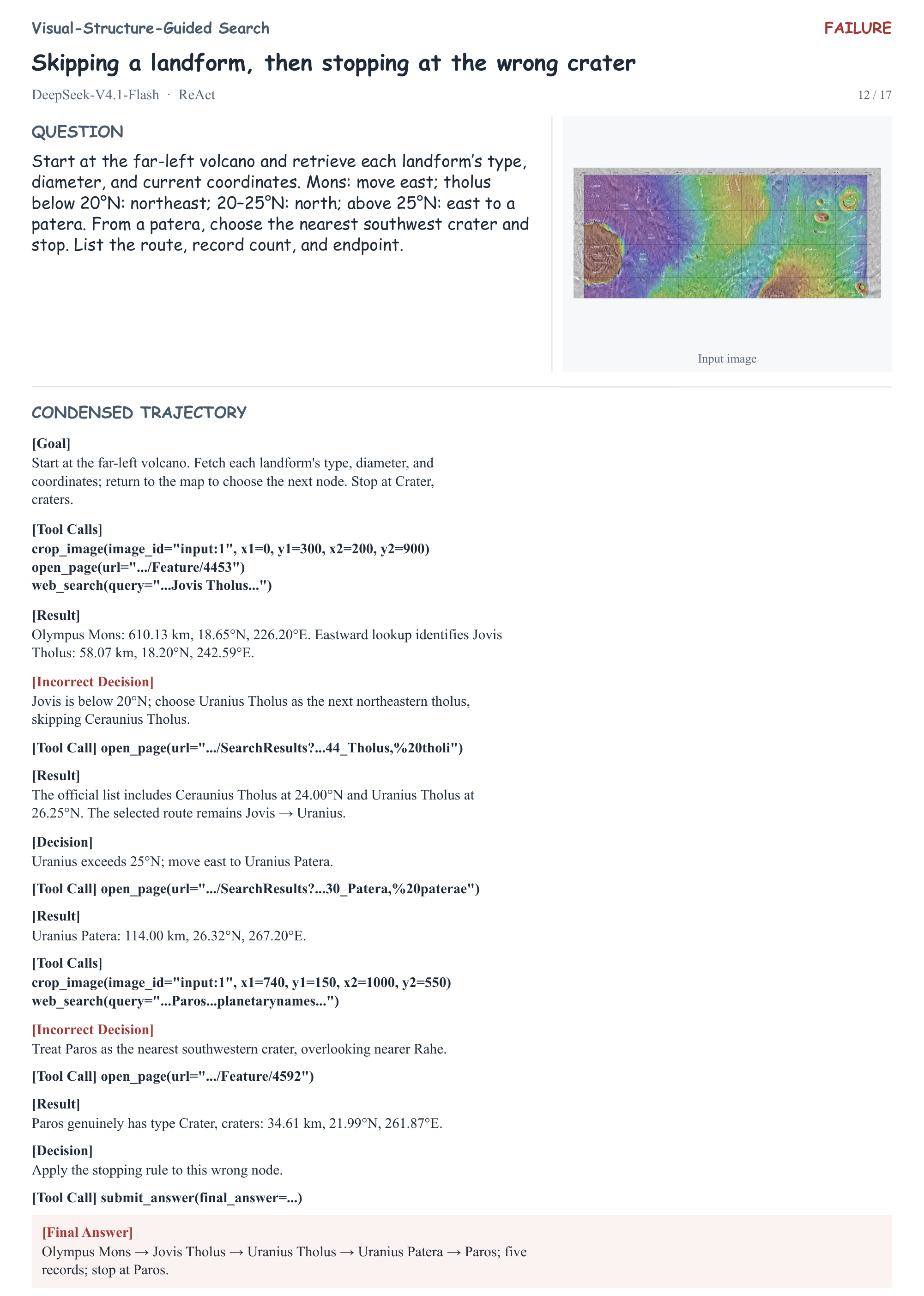}
\caption{A failed Visual-Structure-Guided Search trajectory from DeepSeek-V4.1-Flash (ReAct).}
\label{fig:supp-rec-failure}
\end{figure}

\begin{figure}[p]
\centering
\includegraphics[width=\textwidth,height=0.90\textheight,keepaspectratio]{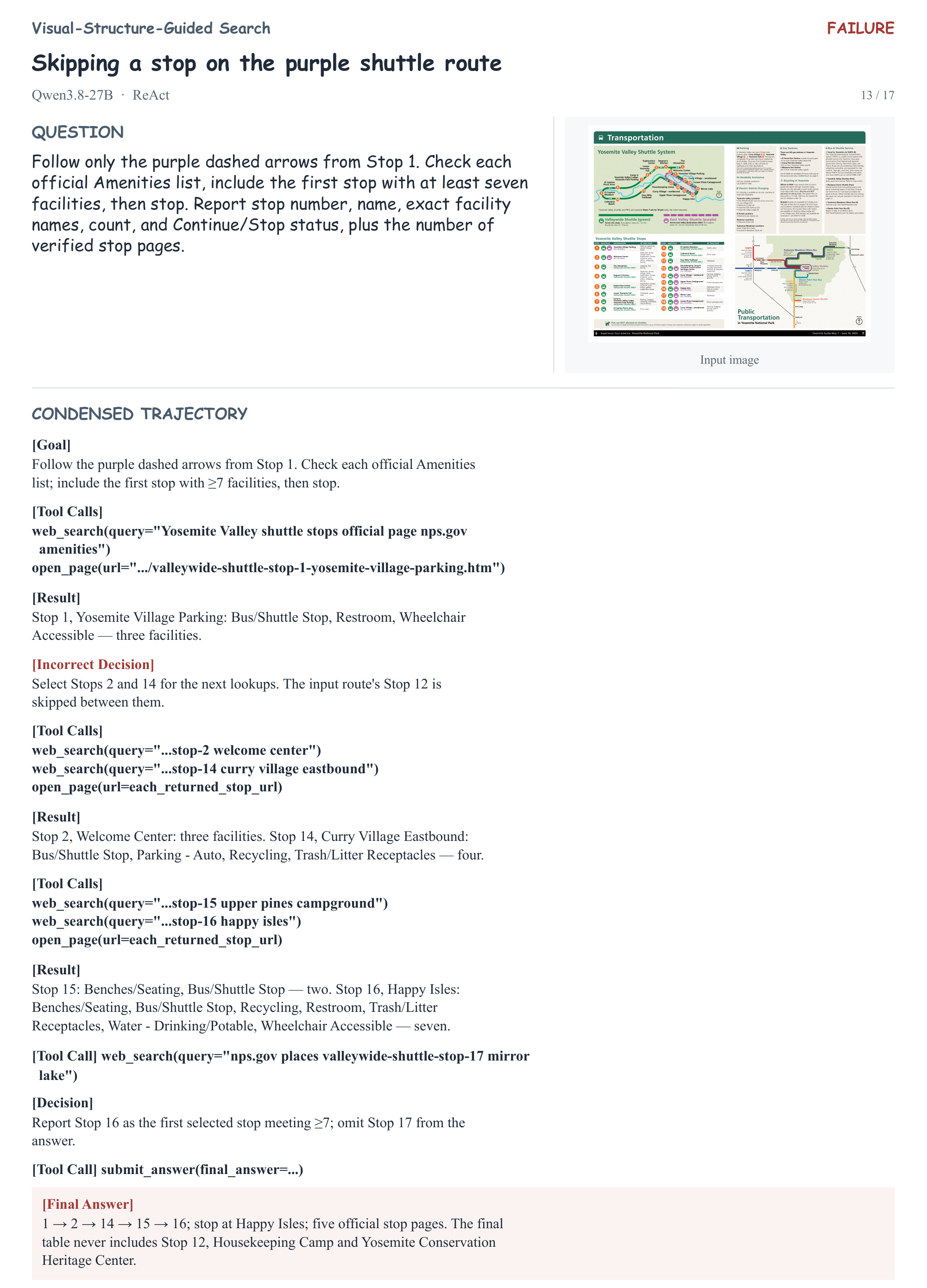}
\caption{A failed Visual-Structure-Guided Search trajectory from Qwen3.8-27B (ReAct).}
\label{fig:supp-rec-failure-2}
\end{figure}

\begin{figure}[p]
\centering
\includegraphics[width=\textwidth,height=0.90\textheight,keepaspectratio]{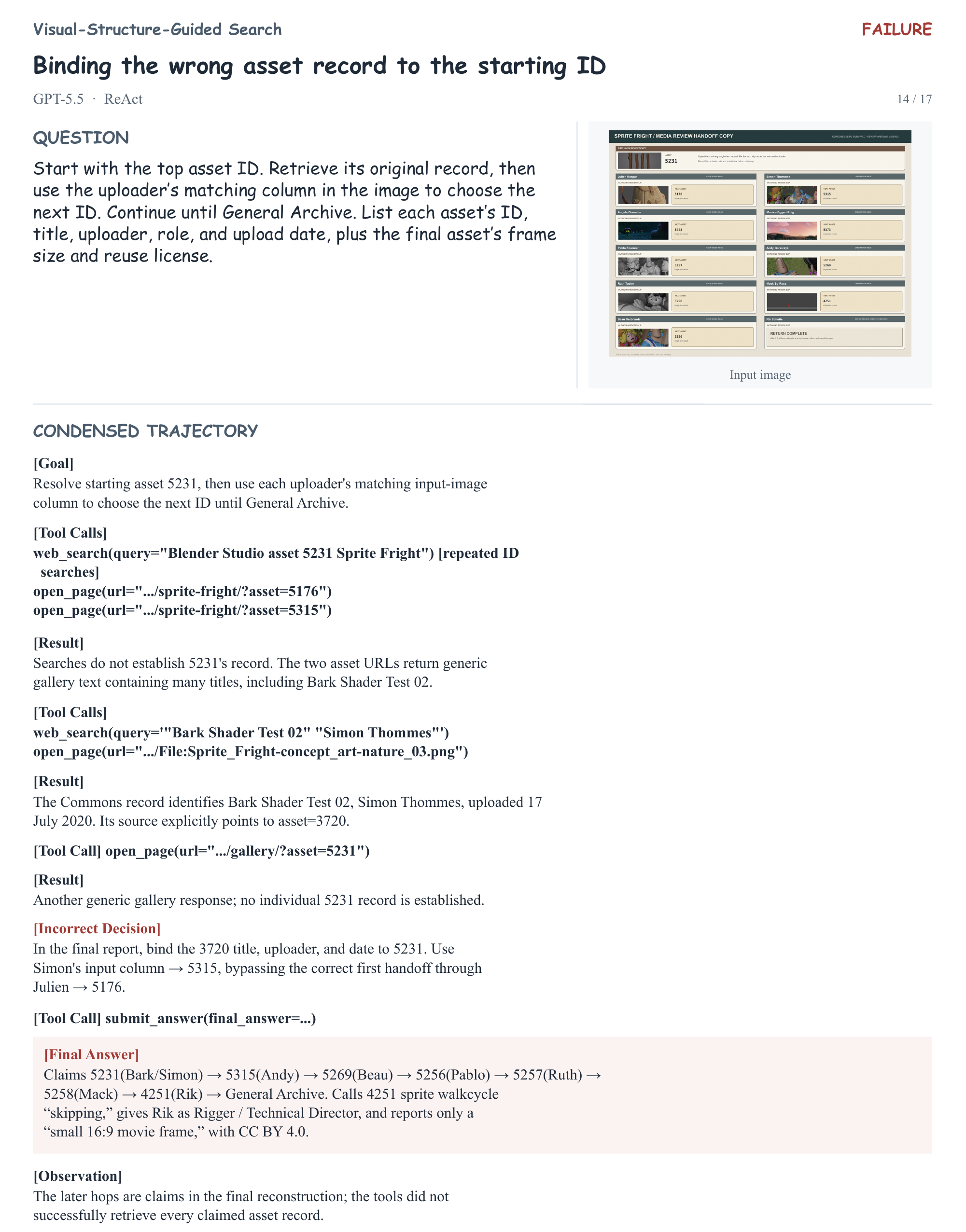}
\caption{A failed Visual-Structure-Guided Search trajectory from GPT-5.5 (ReAct).}
\label{fig:supp-rec-failure-3}
\end{figure}

These tasks combine image order, positions, or connections with external
results to select the next target and determine when to stop. Successful
cases use uploader records to return to the image and select the next asset
(Figure~\ref{fig:supp-rec-success}), and check advisories in order until the
wind threshold is reached (Figure~\ref{fig:supp-rec-success-2}). Failures
include selecting the wrong next landform (Figure~\ref{fig:supp-rec-failure}),
skipping a route stop (Figure~\ref{fig:supp-rec-failure-2}), and linking the
wrong asset record to the starting ID (Figure~\ref{fig:supp-rec-failure-3}).
These errors change the path defined by the image and misdirect later
searches and answers.

\clearpage

\clearpage
\clearpage

\clearpage
\clearpage

\clearpage
\clearpage

\clearpage
\clearpage

\clearpage
\clearpage

\clearpage
\clearpage

\clearpage
\clearpage

\clearpage
\clearpage

\clearpage
\clearpage

\clearpage
\clearpage

\clearpage
\clearpage

\clearpage
\clearpage

\clearpage
\clearpage

\clearpage
\clearpage

\clearpage
\clearpage

\clearpage
\clearpage

\clearpage
\clearpage

\clearpage

\end{document}